\documentclass{article}

\usepackage{iclr2026/iclr2026_conference,times}
\iclrfinalcopy

\usepackage[utf8]{inputenc} % allow utf-8 input
\usepackage[T1]{fontenc}    % use 8-bit T1 fonts
\usepackage{hyperref}       % hyperlinks
\usepackage{url}            % simple URL typesetting
\usepackage{booktabs}       % professional-quality tables
\usepackage{amsfonts}       % blackboard math symbols
\usepackage{nicefrac}       % compact symbols for 1/2, etc.
\usepackage{microtype}      % microtypography
\usepackage[dvipsnames]{xcolor}        % colors

\usepackage{titlesec}
\titlespacing*{\paragraph}{\parindent}{0.25ex}{1ex}
\usepackage{inconsolata}
\usepackage{amsmath}
\usepackage{cleveref}
\usepackage{tikz}
\usepackage{tikz-dependency}
\usepackage{examples-slim}
\usepackage{amsthm}
\usepackage{enumitem}
\usepackage{bm}
\usepackage{subcaption}
\usepackage{multirow}
\usepackage{adjustbox}
\usepackage{minitoc}

\usepackage{tikz}
\usetikzlibrary{positioning, arrows.meta, backgrounds}

\usepackage{dsfont}

\theoremstyle{definition}
\newtheorem{definition}{Definition}[section]

\crefformat{section}{\S#2#1#3}
\crefformat{subsection}{\S#2#1#3}
\crefformat{subsubsection}{\S#2#1#3}
\crefformat{paragraph}{\P#2#1#3}
\crefformat{subparagraph}{\P#2#1#3}
\crefmultiformat{section}{\S#2#1#3}{ and~\S#2#1#3}{, \S#2#1#3}{, and~\S#2#1#3}
\crefmultiformat{subsection}{\S#2#1#3}{ and~\S#2#1#3}{, \S#2#1#3}{, and~\S#2#1#3}
\crefmultiformat{subsubsection}{\S#2#1#3}{ and~\S#2#1#3}{, \S#2#1#3}{, and~\S#2#1#3}
\crefmultiformat{paragraph}{\P\P#2#1#3}{ and~#2#1#3}{, #2#1#3}{, and~#2#1#3}
\crefmultiformat{subparagraph}{\P\P#2#1#3}{ and~#2#1#3}{, #2#1#3}{, and~#2#1#3}
\crefrangeformat{section}{\mbox{\S\S#3#1#4--#5#2#6}}
\crefrangeformat{subsection}{\mbox{\S\S#3#1#4--#5#2#6}}
\crefrangeformat{subsubsection}{\mbox{\S\S#3#1#4--#5#2#6}}
\crefrangeformat{paragraph}{\mbox{\P\P#3#1#4--#5#2#6}}
\crefrangeformat{subparagraph}{\mbox{\P\P#3#1#4--#5#2#6}}
\crefname{part}{Part}{Parts}
\Crefname{part}{Part}{Parts}
\crefname{chapter}{Ch.}{Ch.}
\Crefname{chapter}{Ch.}{Ch.}
\crefname{footnote}{Fn.}{Fn.}
\Crefname{footnote}{Fn.}{Fn.}
\crefname{figure}{Figure}{Figures}
\crefname{table}{Table}{Tables}
\crefname{subfigure}{Figure}{Figures}
\Crefname{subfigure}{Figure}{Figures}
\crefname{appsec}{Appendix}{Appendices}
\Crefname{appsec}{Appendix}{Appendices}
\crefname{algocf}{Algorithm}{Algorithms}
\Crefname{algocf}{Algorithm}{Algorithms}
\crefname{xnumi}{ex.}{exs.}
\Crefname{xnumi}{Ex.}{Exs.}
\crefname{xnumii}{ex.}{exs.}
\Crefname{xnumii}{Ex.}{Exs.}

\renewcommand\cite{\citep}% to get "(Author Year)" with natbib    
\PassOptionsToPackage{breaklinks}{hyperref}
\RequirePackage{hyperref}
\definecolor{darkblue}{rgb}{0, 0, 0.5}
\hypersetup{colorlinks=true, citecolor=darkblue, linkcolor=darkblue, urlcolor=darkblue}

\newcommand{\ourname}{Associative Treecall}
\newcommand{\ourshort}{ATR}

\newcommand{\keytok}{\textcolor{BrickRed}{\textit{key}}}
\newcommand{\valuetok}{\textcolor{RoyalBlue}{\textit{value}}}
\newcommand{\querytok}{\textcolor{ForestGreen}{\textit{query}}}
\newcommand{\answertok}{\textcolor{DarkOrchid}{\textit{answer}}}

\title{Mechanistic evaluation of \\ Transformers and state space models}

\author{%
  Aryaman Arora \quad Neil Rathi \quad Nikil Roashan Selvam \quad Róbert Csordás \\ \textbf{Dan Jurafsky} \quad \textbf{Christopher Potts} \\
  Stanford University \\
  \texttt{\{aryamana,jurafsky,cgpotts\}\@stanford.edu} \\
}

\begin{document}
\doparttoc % Tell to minitoc to generate a toc for the parts
\faketableofcontents % Run a fake tableofcontents command for the partocs

\part{} % Start the document part

\maketitle

\begin{abstract}
State space models (SSMs) for language modelling promise an efficient and performant alternative to quadratic-attention Transformers, yet show variable performance on recalling basic information from the context. While performance on synthetic tasks like Associative Recall (AR) can point to this deficiency, behavioural metrics provide little information as to \textit{why}---on a mechanistic level---certain architectures fail and others succeed.
To address this, we conduct experiments on AR, and find that only Transformers and Based SSM models fully succeed at AR, with Mamba and DeltaNet close behind, while the other SSMs (H3, Hyena) fail. We then use causal interventions to explain why.
We find that Transformers and Based learn to store key--value associations in-context using induction. By contrast, the SSMs seem to compute these associations only at the last state using a single layer. We further investigate the mechanism underlying the success of Mamba, and find novel evidence that Mamba \textit{does} implement induction: not via the SSM, but instead via short convolutions.
Further experiments on a new hierarchical retrieval task, \ourname{} (\ourshort{}), show that all architectures learn the same mechanism as they did for AR. Furthermore, we show that Mamba can learn Attention-like induction on \ourshort{} when short convolutions are removed.
% To extend and deepen these findings, we introduce \ourname{} (\ourshort{}), a synthetic task similar to AR based on PCFG induction. \ourshort\ introduces language-like hierarchical structure into the AR setting.
% We find that all architectures learn the same mechanism as they did for AR, and the same three models succeed at the task.
These results reveal that architectures with similar accuracy may still have substantive differences, motivating the adoption of mechanistic evaluations.
%%%%%%%%%%%%%% TODO: COMMENT BACK %%%%%%%%%%%%%% 
\begin{center}
\small
\raisebox{-0.2\height}{\includegraphics[width=1em,height=1em]{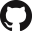}}\hspace{0.5em}\href{https://github.com/aryamanarora/tinylang}{\texttt{github.com/aryamanarora/tinylang}}
\end{center}
% \npr{ATR part should be rewritten prolly}

% Subquadratic attention architectures for language modelling achieve impressive benchmark scores, but fail to outperform quadratic attention at in-context retrieval. While synthetic tasks such as associative recall (AR) have helped elucidate this deficiency at a behavioural level, our understanding of what mechanism quadratic attention learns for such tasks and why efficient alternatives fail is still limited. Therefore, we introduce \ourname{} (\ourshort{}), a synthetic task similar to AR which breaks naïve positional associations and thus enables studying how both association and retrieval challenge language models at small scales when pretraining from scratch. We apply tools from mechanistic interpretability, such as interchange intervention, to uncover the internal mechanisms learned for this task across a variety of architectures. We find that subquadratic attention architectures learn an entirely different algorithm \npr{what algorithm? maybe more explicit. also mention that quad attn learns some sort of `correct' algorithm} for \ourshort{} than attention, explaining poor state-size scaling and lack of positional understanding.
\end{abstract}

\section{Introduction}

Transformers with quadratic attention remain the dominant architecture in language modelling despite numerous proposed efficient alternatives. Most notably, \textbf{state space models} (SSMs) achieve impressive perplexities and benchmark scores \cite[e.g.][]{gu2024mambalineartimesequencemodeling}. Yet, SSMs exhibit deficiencies that benchmarks often fail to capture; for example, they struggle to perform \textbf{retrieval}, i.e. copying from the context \cite{jelassi2024repeat,wen2024rnnstransformersyetkey,waleffe2024empiricalstudymambabasedlanguage,bick2025understandingskillgaprecurrent}.

Controlled synthetic tasks can make these limitations clear by isolating specific capabilities and enabling expressive experimentation at small scales across architectures. Particularly, much work has used the \textbf{associative recall} (AR) task as a testbed for studying in-context retrieval across architectures. In turn, AR has informed the design of novel LM architectures \citep[e.g.~Based;][]{arora2024linear}. % The difficulty of AR can be easily controlled by varying the number of key--value pairs in context; this parametrisation enabled the observation of poor scaling behaviour in SSMs by \citet{arora2024zoology}.

However, in this prior work, performance on synthetic tasks is measured solely via behavioural metrics like task accuracy. This is a missed opportunity: an advantage of these tasks is that they are designed to isolate a \textit{specific behaviour} that implicates a mechanistic solution. For example, LMs should solve AR by storing \keytok{}--\valuetok{} associations in-context at the \valuetok{}.
For Transformer-based LMs \citep{attention}, the mechanism is generally thought to be the \textbf{induction head}. \cite{olsson2022incontextlearninginductionheads,fu2023hungry}.
% \npr{we don't mention induction heads before this, i am skeptical that all readers will know what they are}
We should therefore directly check whether each architecture learns an induction mechanism, as a way of checking that it has learned a robust solution to the task. 
%as part of performance evaluation on AR.
% Direct mechanistic evaluation clarifies fundamental differences between architectures, which can in turn inform new architecture design and allow us to understand model behaviour in more complex settings.

% However, performance on these controlled tasks is still measured purely via behavioural metrics, such as task accuracy.
% Given that synthetic tasks isolate a specific behaviour which implicates a desired mechanistic solution (e.g.~induction heads for AR), why not use tools from mechanistic interpretability to \textit{directly analyse which mechanisms these models are learning}? \npr{probably rephrase what we mean by `correct'}
% \npr{
% However, performance on these controlled tasks is still measured purely via behavioural metrics, such as task accuracy.
% This is a missed opportunity: an advantage of these synthetic tasks is that they are designed to isolate a \textit{specific behaviour} that implicates a mechanistic solution---for example, language models ought to solve AR using induction heads \cite{fu2023hungry}. We ought to directly check whether the desired mechanism is being learned.
% We therefore propose using tools from mechanistic interpretability to directly analyse which mechanisms are present.
% }

Here, we propose using tools from mechanistic interpretability to directly analyse the mechanisms used to solve AR tasks. We use \textbf{causal interventions} \cite{geiger2024causalabstractiontheoreticalfoundation} on model internals to understand how these tasks are learned and implemented across a variety of architectures (\cref{sec:mechanistic}). This allows us to track the emergence (or lack thereof) of the correct association and retrieval mechanisms inside the model, which moves us beyond observed task accuracy and towards a detailed picture of the solutions different architectures learn.

Through comprehensive experiments on AR, we find that all SSMs except Based do not use induction in the traditional sense to implement AR (\cref{sec:ar-exp}); instead, we find that the best-performing SSM, Mamba, relies heavily on the short convolution component in each layer to perform \keytok{}-\valuetok{} association, and fails to learn AR at all without this (\cref{sec:ar-short-conv}). Through layer-internal interventions, we discover that the short convolution component implements an induction-like associative mechanism in Mamba, with the SSM component only being used for retrieval (\cref{sec:ar-mamba-induction}).

To deepen our findings, we introduce \textbf{\ourname{}} (\ourshort{}), a novel retrieval task more similar to real-world natural language retrieval than AR (\cref{sec:atr}). \ourshort{} uses a probabilistic context-free grammar (PCFG) to generate hierarchical data, on which we ask AR-like queries; the key difference from AR is that \keytok{}s and their associated \valuetok{}s need not be adjacent to each other. We find that the same mechanisms are implicated across architectures on ATR as on AR, confirming the generality of our findings (\cref{sec:atr-exp}). However, unlike AR, Mamba does not need short convolutions to learn ATR; instead, mechanistic metrics reveal that Mamba falls back to an Attention-like two-layer induction mechanism when its short convolutions are removed (\cref{sec:atr-mamba}).

\begin{figure}
    \centering
    \includegraphics[width=\linewidth]{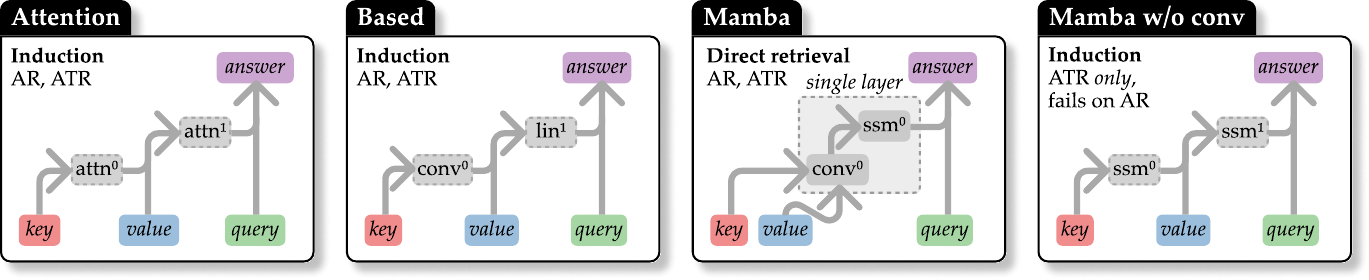}
    \caption{\textbf{Associative mechanisms in Attention, Based, and Mamba}: Key results from applying mechanistic metrics to understand how Attention and SSMs solve AR and our new retrieval task, \ourname{} (\ourshort{}). Attention and Based both implement induction, but with different architectural components (\cref{sec:ar-exp,sec:atr-exp}). Mamba instead uses a single layer for both association and retrieval, but uses short convolution for association just like Based (\cref{sec:ar-short-conv,sec:ar-mamba-induction}). When convolutions are removed, Mamba implements induction, but only on \ourshort{} (\cref{sec:atr-mamba}).}
    \label{fig:overview}
\end{figure}

We offer a framework for better understanding and evaluating task performance via mechanistic interpretability. We demonstrate multiple examples of mechanistic evaluations revealing fundamental differences between architectures that are unknown via behavioural performance, summarised in \cref{fig:overview}. We thus introduce mechanistic metrics as a new tool for architecture analysis.

\section{Related work}
\label{sec:related}

\paragraph{Associative Recall.} Our work relates to a variety of extant synthetic retrieval tasks. The foremost example, Associative Recall (AR),\footnote{Also known as \textit{associative retrieval}, \textit{associative memory}, or \textit{induction}.} is a synthetic task that evaluates in-context retrieval for language model architectures. AR has been used extensively, from early work on recurrent neural networks \citep{graves2014neuralturingmachines,ba2016fast,danihelka2016assoc,zhang2017learningupdateautoassociativememory} to modern SSMs \citep{fu2023hungry,poli2023hyena,lutati-etal-2023-focus,jelassi2024repeat,arora2024zoology,arora2024linear,gu2024mambalineartimesequencemodeling,dao2024transformersssmsgeneralizedmodels,trockman2024mimeticinitializationhelpsstate,liu2024longhornstatespacemodels,okpekperevisiting,li2025cat,wang2025testtimeregressionunifyingframework}.

An AR task consists of a sequence of key--value pairs followed by a single \textit{query} key; the goal is to produce the corresponding value for the given query. For example,
\begin{examples}
\centering
\item \texttt{A 2 \textbf{C 3} F 9 D 1 \textbf{C} $\to$ \textbf{3}} \label{ex:ar}
\end{examples}
Here, the correct next token is \texttt{3}, since it is the value associated with the key \texttt{C} in context.

% A few variants of AR have been proposed in the literature, including Relational AR \citep{le2020self}, AR with rewrites \citep{rodkin2025associativerecurrentmemorytransformer}, and multi-query AR \citep{arora2024zoology}.\footnote{They note that ``[c]ompared to the prior AR formulations, MQAR better captures the persisting quality gaps on synthetic and real world data. However, it is not clear why MQAR elucidates the gap.''}
Despite being synthetic, AR has a direct analogue in natural language: \textit{induction}, referring to in-context copying of sequences \citep{elhage2021mathematical,olsson2022incontextlearninginductionheads}. \citet{arora2024zoology,arora2024linear} show that architecture-level improvements on AR translate directly to induction. 
% Correspondingly, the same mechanism (\textit{induction heads}) in Transformers is implicated.
Association is additionally widely studied in cognitive science as \textit{binding}. Binding in neural networks has been examined by \citet{greff2020bindingproblemartificialneural,kim-schuster-2023-entity,feng2024bind,prakash2024binding,li2025howlanguagemodelstrack,prakash2025languagemodelsuselookbacks}.
% Yet no synthetic language-like binding task exists for studying this mechanism in LMs. Finally, besides AR, formal languages are widely used to design synthetic tasks for language modelling \citep[][\textit{inter alia}]{white-cotterell-2021-examining,valvoda-etal-2022-benchmarking,hahn2023theoryemergentincontextlearning,strobl-etal-2024-formal,allenzhu2024physicslanguagemodels1,akyurek2024incontext,pandey2024gzippredictsdatadependentscaling,lubana2024percolationmodelemergenceanalyzing}.

\paragraph{Mechanistic interpretability.} In order to measure the contribution of individual model components (neurons, layers, etc.)\ to output behaviour, we can apply causal interventions on neural network internals \citep{causalabstraction,geiger2024causalabstractiontheoreticalfoundation}. Informally, the core idea is to overwrite an activation at a specific component using a counterfactual input. If this changes model behaviour, then that component is causally relevant to the mechanism underlying that behaviour.

Some prior work in mechanistic interpretability has studied how some language models solve in-context retrieval tasks like induction and multiple choice question answering \citep{olsson2022incontextlearninginductionheads, lieberum2023does, brinkmann-etal-2024-mechanistic, wiegreffe2025answer, bick2025understandingskillgaprecurrent}, as well as the training dynamics of Transformers on toy tasks using mechanistic metrics \citep{nanda2023progress,reddy2024mechanistic,singh2024what,edelman2024evolution,tigges2024llm,yin2025attentionheadsmatterincontext}. Yet thus far, \textit{architectural comparisons} on synthetic tasks have not made use of causal interventions.

\section{Mechanistic metrics}
\label{sec:mechanistic}

\begin{figure}
    \centering
    \includegraphics[width=0.7\linewidth]{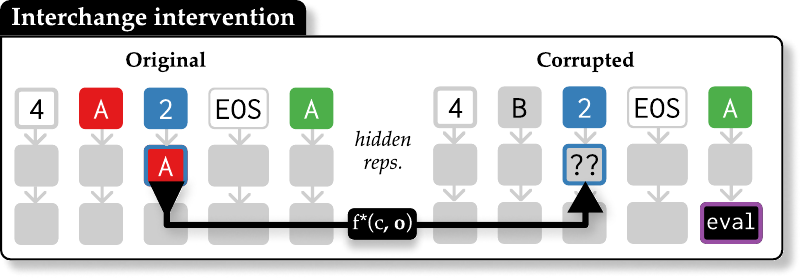}
    \caption{An example interhcange intervention where we corrupt the \keytok{} (\texttt{A}) and attempt to restore it by intervening at the \valuetok{} token in an intermediate representation. We evaluate the downstream effect on the next-token prediction at the \querytok{} (which should ideally predict the \answertok{}).}
    \label{fig:intervention}
\end{figure}

Behavioural metrics provide little information as to \textit{why} certain architectures succeed or fail on tasks of interest. 
% Such metrics are also highly dependent on training hyperparameters. For example, \citet{arora2024zoology} report poor performance of Mamba on AR as difficulty increases, but \citet{okpekperevisiting} find that their LR tuning grid was too sparse; finer tuning of LR on Mamba leads to performance competitive with attention.
Mechanistic metrics, which directly measure how information flows across model components and token positions, can tell us how AR and similar tasks are being solved by different architectures, and thus help us understand failures. We illustrate our approach in \cref{fig:intervention}.

We use interchange interventions \citep{causalabstraction,geiger2024causalabstractiontheoreticalfoundation} to understand and measure how solutions to AR are implemented across architectures. We introduce this operation and define the resulting metrics for our tasks below. Our implementation uses the \texttt{pyvene} library \cite{wu-etal-2024-pyvene}.

\paragraph{Interchange intervention.} Consider a language model $p(\cdot)$ and some input $\mathbf{b}$. We select a component $f$ inside that model which computes some internal representation $f(\mathbf{b})$ during the LM's forward pass. Now, consider a counterfactual input $\mathbf{s}$ which produces a counterfactual representation $f(\mathbf{s})$ when processed by $f$. We want to understand what about the output of $p$ is dependent on $f$. Therefore, we perform an intervention which replaces the output $f(\mathbf{b})$ with that of $f(\mathbf{s})$ during the computation of $p(\mathbf{b})$, with the change propagating downstream. The result is notated $p_{f \gets f^*}(\mathbf{b}, \mathbf{s})$.

\paragraph{Concrete setup for AR and \ourshort{}.} We take $\mathbf{o}$ to be a ground-truth document from our data distribution and $\mathbf{c}$ to be a version of that document with exactly one important token corrupted: the \keytok{} (see \cref{fig:intervention}). This corruption significantly reduces task accuracy by removing information that is necessary to answer the AR query.

For each architecture, we intervene at both the input and output each of the following model components $f$: each layer block, each sequence-mixer (e.g.~Attention blocks in each Transformer layer), and each state-mixer (an MLP, except in Mamba, which lacks this component).\footnote{We are not limited to analysing only these components, however; we do experiment with mixer-internal representations in our analysis.} We measure to what extent the intervention can restore the likelihood of the correct answer to the query, i.e.~we compare \textbf{restored likelihood} $p_{f \gets f^*}(y_{\textrm{true}} \mid \mathbf{c}, \mathbf{o})$ with original likelihood $p(y_\textrm{true} \mid \mathbf{o})$ and corrupted likelihood $p(y_\textrm{true} \mid \mathbf{c})$.

% \paragraph{Metrics.} Given the above three quantities, we compute 
% \textbf{difference}, or how much the likelihood shifts due to the intervention, and
% \textbf{attribution score}, or what proportion of the original likelihood was restored by the intervention:
% \begin{align}
%     % \mathrm{Diff}(f) &= \\
%     \mathrm{Attrib}(f) &= \frac{p_{f \gets f^*}(y_{\textrm{true}} \mid \mathbf{c}, \mathbf{o}) - p(y_\textrm{true} \mid \mathbf{c})}{p(y_\textrm{true} \mid \mathbf{o}) - p(y_\textrm{true} \mid \mathbf{c})}
% \end{align}
% For AR and \ourshort{} in particular, there are two choices for $f$ which help us distinguish the mechanism underlying task success. To check whether induction is the underlying mechanism, we compute metrics for $f$ being the layer $1$ \textit{block input} at the \textit{value} token. Alternatively, we check whether other tokens at layer $1$ block input mediate information flow, indicating some sort of association-less direct retrieval mechanism: the \textit{key}, \textit{query}, and \textit{divider}.

\section{Understanding AR with mechanistic metrics}
\label{sec:experiments}

We now deploy our mechanistic metrics (\cref{sec:mechanistic}) on AR. We follow the methodology outlined in \cref{sec:methodology} to create a variety of datasets and train models with various architectures and hyperparameter configurations. See \cref{sec:additional-exps} for additional experiments not included here.

\subsection{Methodology}
\label{sec:methodology}
\paragraph{Datasets.} We generate synthetic pretraining and evaluation datasets for AR. For each setting, the trainset has $100,032$ examples and the eval/dev sets have $320$ examples. In AR, we use disjoint key and value vocabularies. In each document, we separate the document from the query with a divider token, and provide only a single query.
% For example, an AR document is:
% \begin{examples}
% \centering
% \item \texttt{k123 v2184 \textbf{k6342 v96} k824 v13 k5904 v58437 EOS \underline{\textbf{k6342}} \textbf{v96}} \\
% \label{ex:ar-ours}
% \end{examples}
Further details are in \cref{sec:task-params}.

\begin{figure}
    \centering
    % \begin{subfigure}[t]{0.25\linewidth}
    %     \includegraphics[width=\linewidth]{fig/ar_32.pdf}
    %     \caption{Model dimensionality vs.~accuracy after tuning learning rate for each setting.}
    %     \label{fig:ar-behaviour}
    % \end{subfigure}
    % \hspace{0.03\linewidth}
    % \begin{subfigure}[t]{\linewidth}
        \includegraphics[width=\linewidth]{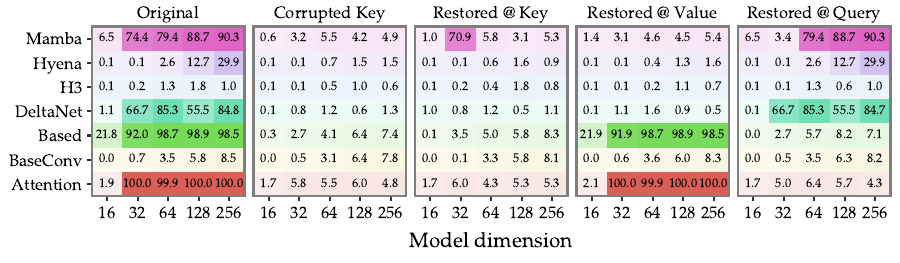}
        % \caption{Attribution score when restoring the key at its \textit{value} (induction) vs.~at the \textit{query}, for all LRs.}
        % \label{fig:ar-interp}
    % \end{subfigure}
    % \begin{subfigure}[t]{0.3\linewidth}
    %     \includegraphics[width=\linewidth]{fig/ar_32_diff_query_item_orig.pdf}
    %     \caption{Attribution score when restoring the key at its \textit{value} (induction) vs.~at the \textit{query}, for all LRs.}
    %     \label{fig:ar-interp}
    % \end{subfigure}
    % \hspace{0.03\linewidth}
    % \begin{subfigure}[t]{0.3\linewidth}
    %     \includegraphics[width=\linewidth]{fig/ar_32_diff_query_item_orig_q.pdf}
    %     \caption{Same as left but with restoring at the \textit{query} (layer 0 direct retrieval) vs.~\textit{key} (layer 1).}
    %     \label{fig:ar-interp-key}
    % \end{subfigure}
    \caption{\textbf{Associative recall}: Likelihood of correct answer without any interventions, after corrupting the key, and after restoring representations at the layer 1 block input with interchange intervention, on AR with vocabulary size $8192$ and key--value count of $32$. SSMs (except for Based) and Transformers learn different mechanisms.}
    \label{fig:ar}
\end{figure}

\paragraph{Models.} We pretrain small models from scratch. We use the exact architecture implementations from the \texttt{zoology}\footnote{\url{https://github.com/HazyResearch/zoology}} library \citep{arora2024linear} as well as the DeltaNet implementation from \texttt{fla}\footnote{\url{https://github.com/fla-org/flash-linear-attention}} \cite{yang2024fla}, except for behaviour-preserving modification of the LM backbone to enable interventions with \texttt{pyvene}\footnote{\url{https://github.com/stanfordnlp/pyvene}} \citep{wu-etal-2024-pyvene} on various model-internal components. The LM backbone for all architectures is the same, with pre-norm blocks of alternating sequence mixers and MLPs (except for Mamba, which has no MLP) followed by LayerNorm at the end.
We experiment with the following architectures: {Attention} \citep{attention}, {BaseConv} \cite{arora2024zoology}, {Based} \cite{arora2024linear}, {DeltaNet} \cite{deltanet}, {H3} \cite{fu2023hungry}, {Hyena} \cite{poli2023hyena}, and {Mamba} \cite{gu2024mambalineartimesequencemodeling}; further details are given in \cref{sec:config}.

\paragraph{Training.} We minimise next-token prediction cross-entropy loss, and mask the loss on all tokens except the \textit{query}. We use the AdamW optimiser with $\beta = (0.9, 0.999), \epsilon = 10^{-8}$ and no weight decay. We warm up the learning rate for the first $10\%$ of training and then follow a cosine decay schedule to $0$ for the remainder of training. We train for $16$ epochs with a batch size of $32$.

Each experiment trains $\approx200$ models over all hyperparameters. Runtime varies from $0.5$ to $5$ hours, depending on hardware, task, and architecture. Overall, we used $<10,000$ GPU-hours in total, on a cluster with various NVIDIA machines (GPU memory $12.3$G to $143.8$G).

\paragraph{Behavioural metrics.} We report behavioural metrics given the model's predicted probabilities over the vocabulary $\hat{\mathbf{y}} \in \mathbb{R}^{\lvert \Sigma \rvert}$ and the index of the single true answer $i$. Our main metric is likelihood of the correct answer: $\hat{\mathbf{y}}_i$. We additionally report accuracy in appendices: $\mathds{1}[\mathrm{arg\,max}(\hat{\mathbf{y}}) = i]$.

\subsection{(Most) SSMs do not solve AR with induction}
\label{sec:ar-exp}

We first run experiments on a standard AR task and show that interchange interventions empirically confirm architectural differences in how AR is solved. We fix the total number of unique keys and values in the vocabulary to be $8192$, and present $32$ key--value pairs in context. As noted above, our trainset includes $100,032$ examples. We vary model dimensionality in $\{16, 32, 64, 128, 256\}$ and sweep LR in the range $[3 \cdot 10^{-5}, 3 \cdot 10^{-2}]$ for each architecture.

\paragraph{Behavioural results.} The leftmost panel of \cref{fig:ar} demonstrates that answer likelihood on AR cleanly separates Attention, which achieves $100\%$ likelihood and accuracy at $d \geq 32$, from nearly all SSMs. Based solves AR near-perfectly with roughly the same dimension-wise scaling curve as Attention, achieving a maximum likelihood of $98.9\%$ (accuracy of $99.06\%$). However, Mamba and DeltaNet are close behind and clearly better than other SSMs at AR, albeit achieving a less-than-perfect $90.3\%$ likelihood ($91.25\%$ accuracy) at $d = 256$.

\paragraph{Mechanistic analysis.} After corrupting the \keytok{}, we restore the original representation at the layer 1 block input at each of various positions (\keytok{}, \valuetok{}, and \querytok{}) and observe the resulting likelihood of the true answer. High restored likelihood at the \valuetok{} token causally indicates that the model is performing \textbf{induction}, whereas \querytok{} indicates \textbf{direct retrieval} by the layer 0 block and \keytok{} indicates the layer 1 block, respectively. We report results on the LR-tuned checkpoints of each architecture at each model dimension.

% We compute $\mathrm{Attrib}$ for layer 1 block input at the \textit{value} token vs.~\textit{query} token for all training runs where $p(y_\textrm{true} \mid \mathbf{o}) - p(y_\textrm{true} \mid \mathbf{c}) > 0.01$.\footnote{We filter in order to discard low-performing and noisy runs.} A high attribution score on the \textit{value} token indicates \textbf{induction} as the underlying mechanism while \textit{query} indicates \textbf{direct retrieval} at the final state, performed in layer 0.

Our results in \cref{fig:ar} cleanly separate Attention and Based, which only perform induction, from other SSMs, which either perform direct retrieval (Mamba, DeltaNet, Hyena) or fail to learn the task (H3, BaseConv).
% While only a single BaseConv checkpoint passes our filter, it has the greatest attribution score on the \textit{value}, indicating an induction mechanism.
SSMs perform direct retrieval at varying layers: the best-performing SSMs almost entirely perform direct retrieval at layer 0 via the \querytok{} token, with only one Mamba checkpoint ($d=32$) delays retrieval to layer 1. \citet{jelassi2024repeat} show that direct retrieval in SSMs has asymptotically worse capacity than the induction solution, and this is reflected in performance on AR.

\subsection{Short convolutions enable AR in Mamba and Based}
\label{sec:ar-short-conv}

\begin{figure}
    \centering
    % \begin{subfigure}[t]{0.3\linewidth}
    %     \includegraphics[width=\linewidth]{fig/mamba_conv_size.pdf}
    %     \caption{Accuracy on AR (length $32$) for \textbf{Mamba} when varying the kernel size of the short convolution.}
    %     \label{fig:mamba-behaviour}
    % \end{subfigure}
    % \hfill
    \begin{subfigure}[t]{0.67\linewidth}
        \centering
        \includegraphics[width=0.8\linewidth]{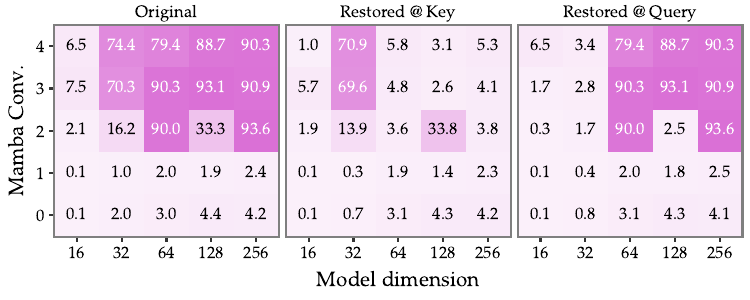}
        \caption{Original and intervention restoration scores for \textbf{Mamba} when varying conv.~kernel size; Restoring at the \valuetok{} is not shown since none of the models benefited from it.}
        \label{fig:mamba-interp}
    \end{subfigure}
    \hfill
    \begin{subfigure}[t]{0.3\linewidth}
        \centering
        \includegraphics[width=0.8\linewidth]{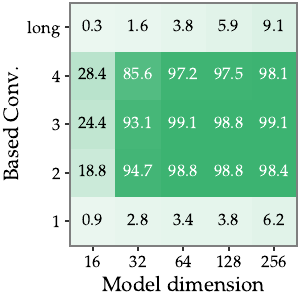}
        \caption{Accuracy on AR for \textbf{Based} when varying conv.~kernel size or using implicit long conv.}
        \label{fig:based-behaviour}
    \end{subfigure}
    \caption{\textbf{Ablating short convolution in Mamba and Based}: Accuracy and interchange intervention results when ablating parameters of the short convolution component in Mamba and Based. Conv.~size less than $2$, using long conv., or no conv., all lead to near-zero performance in both models.}
    \label{fig:short-conv}
\end{figure}

We have observed that Attention, Based, and Mamba are the highest-performing architectures on AR. However, their underlying mechanisms differ: Attention and Based learn \textbf{induction}, a 2-layer mechanism which stores \keytok{}--\valuetok{} associations at the \valuetok{} token as an intermediate step, whereas Mamba uses \textbf{direct retrieval}, a 1-layer mechanism which writes the association to the \querytok{} token.

Importantly, Based and Mamba share a key architectural component:~\textbf{short convolutions}. Based is a hybrid model with alternating short convolution and linear attention layers, while Mamba applies a short convolution to the layer input before each SSM block. We hypothesise that this architectural component is necessary\footnote{Since Hyena also has a short convolution, this may not be \textit{sufficient} for good performance on association.} for learning AR when using a subquadratic sequence mixer. We conduct experiments on AR where we shorten the convolution kernel size in Mamba (from the default $d_{\mathrm{conv}} = 4$ to $\{3, 2, 1\}$, and deleting it) and replace the Based short convolution with implicitly-parametrised long convolution \citep{poli2023hyena}.

\paragraph{Results.} We report results of our ablations in \cref{fig:short-conv}. On Mamba (\cref{fig:mamba-interp}), we find a step change in task accuracy when increasing $d_{\mathrm{conv}}$ from $1$ to $2$, which introduces previous token information and thus enables AR. Without short convolution, Mamba fails to learn AR. Also, no Mamba checkpoints store the \keytok{} at the \valuetok{} (i.e.~induction), so direct retrieval persists. Finally, besides $d_{\mathrm{conv}} < 2$ like Mamba, implicit long convolution in Based also significantly harms AR performance (\cref{fig:based-behaviour}). We thus conclude that short convolutions are necessary for performing AR in Mamba and Based.

\subsection{Actually, Mamba \textit{does} do induction}
\label{sec:ar-mamba-induction}

\begin{figure}
    \centering
    \includegraphics[width=\linewidth]{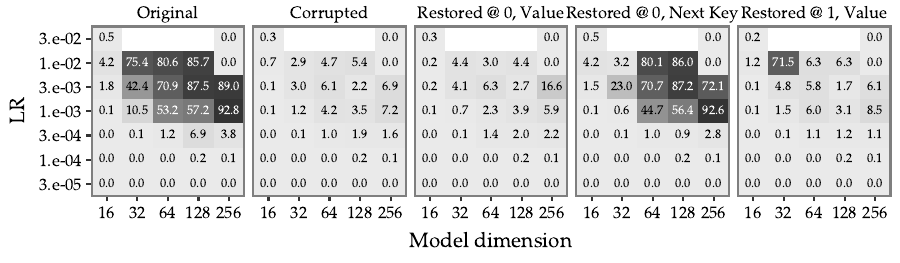}
    \caption{\textbf{Interventions on short convolutions in Mamba}: Likelihood of the correct answer for all Mamba checkpoints (varying model dimension and LR) on AR; (from left to right) original performance, after corrupting the \keytok{}, restoring at the layer 0 short conv output at the \valuetok{}, same but at the next \keytok{}, and same but at the layer 1 \valuetok{} token. (Some runs failed due to high LR.)}
    \label{fig:mamba}
\end{figure}

We have established that \textit{(a)} Attention and Based perform AR via induction, whereas Mamba and other SSMs use direct retrieval, and \textit{(b)} Mamba and Based require short convolutions to succeed at AR. However, we have not elucidated what `direct retrieval' is; its only apparent distinction from induction is that it is implemented using a single layer, but what is happening in that layer is unclear.

We have already observed in Based that the association step of AR can be implemented with a short convolution, and then retrieval can be handled by a component with a longer receptive field (linear attention). But \cref{sec:ar-exp} shows that this results in an induction mechanism in Based: short convolution, just like Attention, performs association by moving information about the key to its associated value. Might short convolution in Mamba serve the same role?

\paragraph{Methodology.} We use mechanistic metrics to analyse each layer's short convolution components in all of our Mamba checkpoints (sweeping model dimension and LR). To enable Mamba-internal interventions, we use a native PyTorch implementation of Mamba and load in the weights from our hardware-optimised training runs; we confirm that unintervened performance is unchanged. As in \cref{sec:ar-exp}, we corrupt the \keytok{}, and attempt to restore \keytok{} information by intervening at the hidden states outputted by the short convolution component in each layer. We report the likelihood of the correct answer after intervention, averaged over $64$ inputs.

\paragraph{Results.} Our results in \cref{fig:mamba} show that short convolution does move the \keytok{} information; however, only a single checkpoint ($d=32$, $\mathrm{LR}=10^{-2}$) actually moves \keytok{} information to the \valuetok{} (as expected in induction, cf.~Attention), and that too in layer 1; instead, we observe in the remaining $9$ checkpoints with non-negligible performance, the information is moved to the \textit{\textbf{next} \keytok{}} by the layer 0 short convolution.
% Additional results in \cref{sec:mamba-more} show that corrupting the \valuetok{} has the same effect. 
While moving both the key and value to the following key is unusual, it still satisfies our two-stage definition of induction, and thus we claim that Mamba too performs induction.

\section{Mechanistic findings on AR generalise to a new retrieval task}
\label{sec:atr}

We now examine whether our results generalise to a novel retrieval task: \textbf{\ourname{}} (\ourshort{}). \ourshort{} is more similar to real-world natural language retrieval than AR because it uses a probabilistic context-free grammar (PCFG) to generate hierarchical data, on which we ask AR-like queries. Since \keytok{} and \valuetok{} need not be adjacent to each other, \ourshort{} requires a non-positional associative mechanism, which may challenge architectures that are designed for AR. 
% Interestingly, we observe the same mechanisms are implicated across architectures on ATR as on AR, indicating that association mechanisms are not task-dependent.
% We only perform experiments on a single setting of ATR in this section; additional settings are studied in \cref{sec:additional-exps}.

\subsection{Associative Treecall (ATR)}

\begin{figure}
    \centering
    \includegraphics[width=0.8\linewidth]{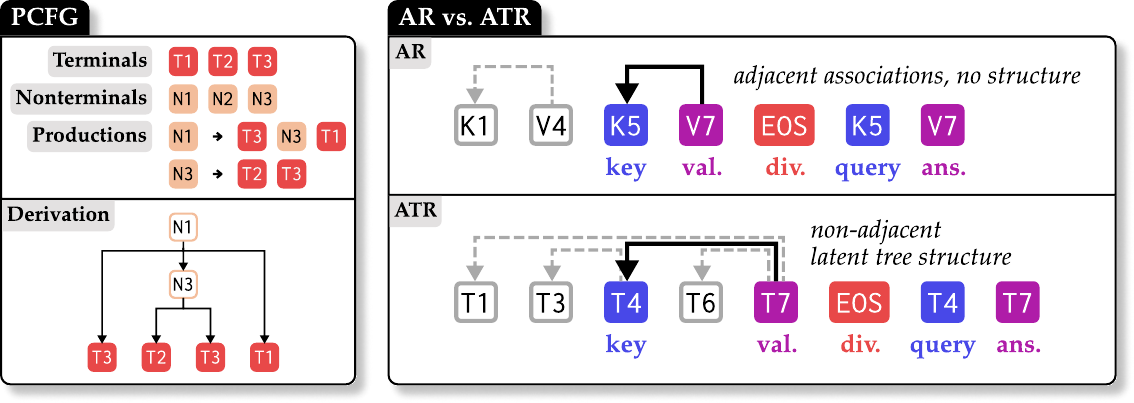}
    \caption{\textbf{PCFG}: An illustrative example of a PCFG and its components, with an example derivation (with final string) below. \textbf{AR vs.~\ourshort{}}: Comparing AR and \ourshort{} using example documents; both tasks provide a document with key--value associations in-context and ask a query about one such association. However, associations in \ourshort{} need not involve adjacent tokens and are tree-structured.}
    \label{fig:atr-ex}
\end{figure}

Since a standard AR document (\cref{ex:ar}) consists of \textit{adjacent} \keytok{}--\valuetok{} pairs, one can associate each \keytok{} with its corresponding \valuetok{} solely using relative position. Yet many natural language retrieval tasks require association over latent hierarchical structure. For example:
\begin{examples}
\centering
\item \textit{\underline{John} had \underline{chicken} and Mary had pork. The \underline{chicken} was eaten by $\to$ \underline{John}}
\end{examples}
Answering this query requires associating \textit{John} with \textit{chicken} and \textit{Mary} with \textit{pork}, and then retrieving the appropriate association for \textit{John}. A solution employing relative positional association would not robust to the possible range of variation (\textit{\underline{John} had some \underline{chicken}}, \textit{\underline{John} decided to have \underline{chicken}}, etc.).

% The purpose of \ourshort{} is to enable studying how architectures implement hierarchical binding as found in the natural language example above, and answer if those solutions are related to those for AR.
% However, no synthetic analogue of this task exists to isolate this mechanism and enable direct comparison to AR. Furthermore, the exact nature of binding in LLMs is not understood.
% \citet{feng2024bind} propose two mechanisms, binding IDs and direct binding, the latter of which is similar to induction and AR. 
% \ourshort{} thus allows us to study how different architectures implement binding compared to simple AR.

An \ourshort{} corpus is drawn from a synthetic probabilistic context-free grammar (PCFG) whose parameters we set. Each document consists of a string sampled from the PCFG, with latent structure made up of \textbf{parent--child} relations between symbols, followed by a divider token (\texttt{EOS}) and a query about one such relation. The PCFG has one special property which establishes the parent--child relationships: for the right-hand side of each production rule, the rightmost symbol is always a terminal, and is the \textit{parent} of the symbols created by this production. We sample strings by selecting an iid nonterminal and recursively applying production rules according to the PCFG distribution. We show an example in \cref{fig:atr-ex} and formalise definitions in \cref{sec:formal}.
Since the number of tokens separating parents and their children may vary, \ourshort{} cannot be solved by a positional associative mechanism.

\paragraph{Setup.} For each experiment, we generate a single PCFG to use across all models to ensure fair comparisons, with parameters in Table~\ref{tab:pcfg-params}. We also reject any samples that have more than $1024$ symbols, which only affects the sampling distribution for the most complex PCFGs we use. We follow \cref{sec:methodology} except we train for $32$ epochs.

Each PCFG sample of length $n$ provides us with a set of $n - 1$ eligible parent--child queries (i.e. a tree with $n-1$ edges). However, terminals may occur multiple times, so a query about a specific symbol may present ambiguity; thus, when presenting a query we consider it to \textit{only} refer to the rightmost instance of that symbol.
% Therefore, the maximum number of eligible queries over all samples is $\min(n - 1, \lvert \Sigma \rvert)$. 
To minimise the ability to heuristically guess, we inversely weight parent--child pairs by the parent's child count when sampling queries.

\subsection{Per-architecture mechanisms are similar between \ourshort{} and AR}
\label{sec:atr-exp}

We consider a simple setting to study models on \ourshort{}: the maximum number of symbols on the right-hand side of any rule $L_\textrm{max} = 5$ and vocabulary size $\lvert\Sigma\rvert = 20$. The remaining parameters are set in accordance with \cref{sec:task-params}; see \cref{tab:pcfg-params} for definitions. We sweep the same model dimensionalities as in \cref{sec:ar-exp}, and a smaller learning rate range of $[3\cdot10^{-5}, 3\cdot10^{-3}]$.

% We consider three initial settings to study models on \ourshort{}, over all combinations of $L_\textrm{max} = \{5, 10\}$ and $\lvert\Sigma\rvert = \{20, 8192\}$ except where both values are large, due to memory constraints. We keep all other parameters fixed with settings given in \cref{sec:task-params}. Varying $L_\textrm{max}$ controls the possible distances between keys and values in the PCFG sample without affecting other properties that play a role in task difficulty (e.g.~depth). Varying $\lvert\Sigma\rvert$ stresses the state capacity, since more key--value pairs must be tracked, without affecting syntactic complexity. We sweep the same model dimensionalities as in \cref{sec:ar-exp}, and a smaller learning rate range of $[3\cdot10^{-5}, 3\cdot10^{-3}]$.

\begin{figure}
    \centering
    % \begin{subfigure}[t]{0.25\linewidth}
    %     \includegraphics[width=\linewidth]{fig/ar_32.pdf}
    %     \caption{Model dimensionality vs.~accuracy after tuning learning rate for each setting.}
    %     \label{fig:ar-behaviour}
    % \end{subfigure}
    % \hspace{0.03\linewidth}
    % \begin{subfigure}[t]{\linewidth}
        \includegraphics[width=\linewidth]{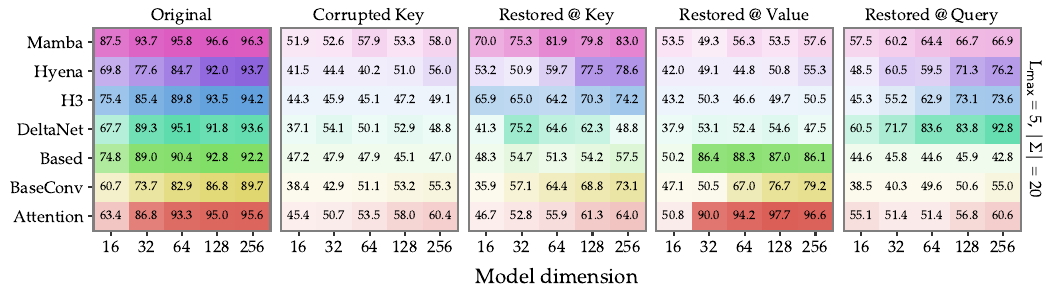}
        % \caption{Attribution score when restoring the key at its \textit{value} (induction) vs.~at the \textit{query}, for all LRs.}
        % \label{fig:ar-interp}
    % \end{subfigure}
    % \begin{subfigure}[t]{0.3\linewidth}
    %     \includegraphics[width=\linewidth]{fig/ar_32_diff_query_item_orig.pdf}
    %     \caption{Attribution score when restoring the key at its \textit{value} (induction) vs.~at the \textit{query}, for all LRs.}
    %     \label{fig:ar-interp}
    % \end{subfigure}
    % \hspace{0.03\linewidth}
    % \begin{subfigure}[t]{0.3\linewidth}
    %     \includegraphics[width=\linewidth]{fig/ar_32_diff_query_item_orig_q.pdf}
    %     \caption{Same as left but with restoring at the \textit{query} (layer 0 direct retrieval) vs.~\textit{key} (layer 1).}
    %     \label{fig:ar-interp-key}
    % \end{subfigure}
    \caption{\textbf{\ourname}: Likelihood of correct answer without any interventions (leftmost), after corrupting the key (second), and after restoring representations at the layer 1 block input (at \keytok{}, \valuetok{}, or \querytok{}) with interchange intervention. The pattern from AR largely holds.}
    \label{fig:atr}
\end{figure}

\paragraph{Behavioural results.} We report results in the leftmost panel of \cref{fig:atr}. Mamba is the best-performing architecture at ATR; it matches or outperforms all other architectures at all model dimensions. This is particularly surprising because longer production rules imply greater positional variation between keys and values, which ought to stress AR-optimised SSM designs like Mamba.

Unlike AR, the baseline performance on \ourshort{} after corrupting the key is well above $0$. This is because some \keytok{}--\valuetok{} pairings are more likely than others due to the underlying PCFG, enabling memorisation (unlike AR where all pairings are equally likely).

\paragraph{Mechanistic analysis.} We conduct the same analysis as for AR. We recover the same overall trends but with greater inter-architecture variance: \cref{fig:atr} shows that Attention, Based, and BaseConv all primarily learn induction mechanisms with performance largely restored by \valuetok{} interventions, whereas the remaining SSMs perform direct retrieval as on AR, with performance being restored by intervening on the \keytok{} (indicating direct retrieval by the layer 1 sequence mixer) or the \querytok{} (indicating the same but by layer 0).

% Intriguingly, \cref{fig:atr} shows that different SSMs form different strategies across task difficulties; in particular, all direct-retrieval SSMs favour earlier retrieval when $\mathrm{L}_{\text{max}}$ is large. Regardless, the same general tendency from AR recurs: SSMs besides Based and BaseConv do not perform induction, but Mamba (and to some extent, DeltaNet) is still highly performant. Strikingly, as the next section shows, Mamba also achieves high generalization performance on ATR.

\subsection{Mamba can learn \ourshort{} without short convolutions}
\label{sec:atr-mamba}

\begin{figure}
    \centering
    \includegraphics[width=\linewidth]{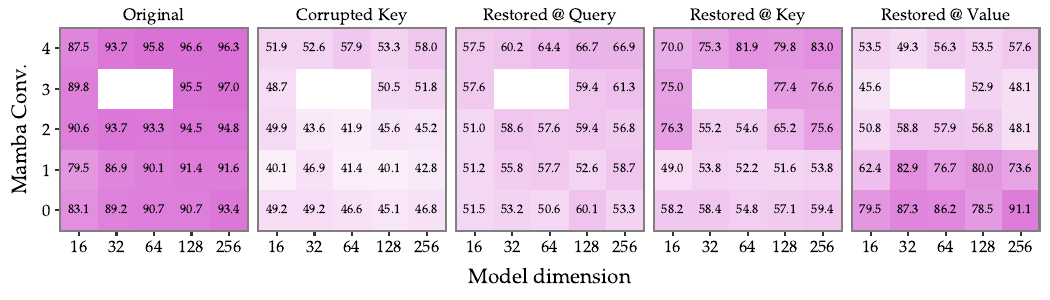}
    \caption{\textbf{Ablating short convolution in Mamba on ATR}: Original, corrupted \keytok{}, and restored interventions at layer 1 input at the \querytok{}, \keytok{}, and \valuetok{} in Mamba, when varying model dimensionality and conv.~size. (Some checkpoints failed to train.)}
    \label{fig:mamba-atr}
\end{figure}

Given \ourshort{} and AR show similar patterns in mechanistic implementation across architectures, we now ask whether Mamba's reliance on short convolution holds on this new task. Since \ourshort{} may have longer and more variable distances between the \keytok{} and \valuetok{} than AR, short convolution (with a short receptive field that applies a fixed input-unaware transformation) may not be the appropriate operator for performing association.

\paragraph{Methodology.} We repeat the experiments in \cref{sec:ar-short-conv} on ATR, training Mamba with varying conv.~size or without conv. We apply mechanistic metrics to the best-performing checkpoints of each dimensionality and conv.~setting.

\paragraph{Results.} We report results in \cref{fig:mamba-atr}. Unlike in AR, we find that Mamba still learns the task even without short convolutions (but not as well as with them). E.g.~at $d=256$, Mamba without short convolutions ($93.4\%$) still outperforms Based ($92.2\%$) and BaseConv ($89.7\%$). The more surprising result is via mechanistic metrics: without short convolutions or with $d_\text{conv}=1$, Mamba's performance after corruption can be restored primarily by intervening at the \valuetok{}, i.e. \textbf{Mamba without short convolutions learns two-layer induction on ATR}. We have thus shown that Mamba can indeed learn an induction-like mechanism to solve associative recall tasks, albeit using different components than Attention and only on certain tasks (\ourshort{}, but not AR). This may explain other induction-like mechanisms observed in Mamba trained on natural language, e.g.~\citet{bick2025understandingskillgaprecurrent}, and highlights a divergence between AR and real-world retrieval.

\section{Discussion}

\paragraph{Why mechanistic evaluations over behavioural metrics?} Architectural advances on language modelling are largely uncovered and presented in an empirical manner (e.g.~\citealp{shazeer2020gluvariantsimprovetransformer}); beyond intuition, we have little justification as to \textit{why} a modification or innovation improves model performance. Synthetic tasks already inform progress on architecture design (such as SSMs), but treating such tasks as just a behavioural evaluation discards useful signal; control over task parameters presents an opportunity to explain performance using interpretability.

\paragraph{ATR indicates induction is highly general.} We introduced ATR to break the naïve key--value adjacency of AR, and see whether general mechanisms underlying association still emerge across architectures. We find the same induction mechanism, where the association is computed and stored at the value before retrieval, in Attention and Based for both tasks, as well as in Mamba without convs. While \citet{olsson2022incontextlearninginductionheads} and later works define induction on adjacent tokens, ATR is evidence that a \textit{position-independent} and generalising (\cref{sec:generalise}) notion of association is worth studying.
% Further investigation of ATR (e.g.~multi-hop queries) is necessary to understand the limits of induction.

\paragraph{Short convolutions are key to association in SSMs.} We showed that Mamba and Based rely on short convolutions to learn how to associate keys and values on AR and ATR. Several earlier works point to the importance of short convolution: \citet{arora2024linear} empirically show its utility on AR, \citet{zeyuan} introduce a short convolution component (Canon) in various architectures,\citet{olsson2022incontextlearninginductionheads} show that 1-layer attention can learn induction if augmented with a length-2 convolution; also \citet{liu2024short,dolga2024lattelatentattentionlinear,fu2023hungry,poli2023hyena}.
%\footnote{We thank \href{https://x.com/dhruv31415/status/1922346462542725614}{Dhruv Pai} for bringing this to our attention.}

\paragraph{Limitations.}\label{sec:limitations}
While we proposed mechanistic evaluations as a new tool, behavioural metrics like accuracy are still needed to properly contextualise results.
% Additionally, here we did not perform mechanistic evaluation of subcomponents of sequence mixers (e.g.~the selective SSM component within Mamba), due to implementation difficulties when applying interventions within hardware-optimised operators, which are inaccessible via PyTorch hooks.
Additionally, we focus on synthetic tasks throughout this work; extending our analyses to real-world models would help paint a more complete picture of the differences in capabilities and mechanisms of architectures on real-world tasks.

\section{Conclusion}
In this work, we introduce mechanistic evaluations as a powerful framework for comparing model architectures. This approach goes beyond high-level behavioural metrics, revealing substantive differences between architectures. Through analysis of synthetic in-context retrieval tasks, we uncover the underlying mechanisms that explain the success and failure points of various architectures. Mechanistic evaluations thus provide a useful tool for architecture design and analysis, as well as a new opportunity for interpretability research to open the blackbox of progress in AI.

\section*{Reproducibility statement}

We provide our anonymised source code for reproducing all experiments at \url{https://anonymous.4open.science/r/tinylang-1061/}. We have specified the hyperparameters for training and evaluation as well as compute usage in \cref{sec:task-params,sec:methodology}. We largely use open-source implementations of models from \texttt{zoology}\footnote{\url{https://github.com/HazyResearch/zoology}} \citep{arora2024linear} as well as the DeltaNet implementation from \texttt{fla}\footnote{\url{https://github.com/fla-org/flash-linear-attention}}, with minor modifications for our mechanistic analysis with the open-source library \texttt{pyvene}\footnote{\url{https://github.com/stanfordnlp/pyvene}} \cite{wu-etal-2024-pyvene}.

% Through thorough analysis on synthetic in-context retrieval tasks---namely AR, and our novel hierarchical variant ATR---we uncover the underlying mechanisms that explain the ability or inability (overselling?) of certain model architectures to succeed at these tasks. Mechanistic evaluations are a useful tool for architecture design and analysis, as well as a new opportunity for interpretability research to open the blackbox of progress in AI.

\section*{Acknowledgements}
We would especially like to thank Zhengxuan Wu, Qinan Yu, Atticus Geiger, and all other attendees of the \texttt{\#weekly-interp-meeting} at Stanford who gave feedback on an early version of this project. We also thank Yanzhe `Sanju' Zhang, Harshit Joshi, Rohan Pandey, Justus Mattern, Ken Ziyu Liu, Julie Kallini, Chenglei Si, Bradley Brown, Jordan Juravsky, Arjun Vikram, and Christine Ye for helpful discussion and feedback at various stages of the project.

This research is supported in part by grants from Google and Open Philanthropy.

% \bibliography{iclr2026_conference}
\bibliography{custom}

@article{graves2014neuralturingmachines,
      title={Neural Turing Machines}, 
      author={Alex Graves and Greg Wayne and Ivo Danihelka},
      year={2014},
      eprint={1410.5401},
      journal={arXiv:1410.5401},
      primaryClass={cs.NE},
      url={https://arxiv.org/abs/1410.5401}, 
}

@inproceedings{ba2016fast,
  author       = {Jimmy Ba and
                  Geoffrey E. Hinton and
                  Volodymyr Mnih and
                  Joel Z. Leibo and
                  Catalin Ionescu},
  editor       = {Daniel D. Lee and
                  Masashi Sugiyama and
                  Ulrike von Luxburg and
                  Isabelle Guyon and
                  Roman Garnett},
  title        = {Using Fast Weights to Attend to the Recent Past},
  booktitle    = {Advances in Neural Information Processing Systems 29: Annual Conference
                  on Neural Information Processing Systems 2016, December 5-10, 2016,
                  Barcelona, Spain},
  pages        = {4331--4339},
  year         = {2016},
  url          = {https://proceedings.neurips.cc/paper/2016/hash/9f44e956e3a2b7b5598c625fcc802c36-Abstract.html},
  bibsource    = {dblp computer science bibliography, https://dblp.org}
}

@article{zhang2017learningupdateautoassociativememory,
      title={Learning to update Auto-associative Memory in Recurrent Neural Networks for Improving Sequence Memorization}, 
      author={Wei Zhang and Bowen Zhou},
      year={2017},
      journal={arXiv:1709.06493},
      archivePrefix={arXiv},
      primaryClass={cs.AI},
      url={https://arxiv.org/abs/1709.06493}, 
}

@inproceedings{danihelka2016assoc,
  author       = {Ivo Danihelka and
                  Greg Wayne and
                  Benigno Uria and
                  Nal Kalchbrenner and
                  Alex Graves},
  editor       = {Maria{-}Florina Balcan and
                  Kilian Q. Weinberger},
  title        = {Associative Long Short-Term Memory},
  booktitle    = {Proceedings of the 33nd International Conference on Machine Learning,
                  {ICML} 2016, New York City, NY, USA, June 19-24, 2016},
  series       = {{JMLR} Workshop and Conference Proceedings},
  volume       = {48},
  pages        = {1986--1994},
  publisher    = {JMLR.org},
  year         = {2016},
  url          = {http://proceedings.mlr.press/v48/danihelka16.html},
  bibsource    = {dblp computer science bibliography, https://dblp.org}
}

@inproceedings{arora2024zoology,
  author       = {Simran Arora and
                  Sabri Eyuboglu and
                  Aman Timalsina and
                  Isys Johnson and
                  Michael Poli and
                  James Zou and
                  Atri Rudra and
                  Christopher R{\'{e}}},
  title        = {Zoology: Measuring and Improving Recall in Efficient Language Models},
  booktitle    = {The Twelfth International Conference on Learning Representations,
                  {ICLR} 2024, Vienna, Austria, May 7-11, 2024},
  publisher    = {OpenReview.net},
  year         = {2024},
  url          = {https://openreview.net/forum?id=LY3ukUANko},
  bibsource    = {dblp computer science bibliography, https://dblp.org}
}

@article{olsson2022incontextlearninginductionheads,
      title={In-context Learning and Induction Heads}, 
      author={Catherine Olsson and Nelson Elhage and Neel Nanda and Nicholas Joseph and Nova DasSarma and Tom Henighan and Ben Mann and Amanda Askell and Yuntao Bai and Anna Chen and Tom Conerly and Dawn Drain and Deep Ganguli and Zac Hatfield-Dodds and Danny Hernandez and Scott Johnston and Andy Jones and Jackson Kernion and Liane Lovitt and Kamal Ndousse and Dario Amodei and Tom Brown and Jack Clark and Jared Kaplan and Sam McCandlish and Chris Olah},
      year={2022},
      journal={arXiv:2209.11895},
      archivePrefix={arXiv},
      primaryClass={cs.LG},
      url={https://arxiv.org/abs/2209.11895}, 
}

@inproceedings{arora2024linear,
  author       = {Simran Arora and
                  Sabri Eyuboglu and
                  Michael Zhang and
                  Aman Timalsina and
                  Silas Alberti and
                  James Zou and
                  Atri Rudra and
                  Christopher R{\'{e}}},
  title        = {Simple linear attention language models balance the recall-throughput
                  tradeoff},
  booktitle    = {Forty-first International Conference on Machine Learning, {ICML} 2024,
                  Vienna, Austria, July 21-27, 2024},
  publisher    = {OpenReview.net},
  year         = {2024},
  url          = {https://openreview.net/forum?id=e93ffDcpH3},
  bibsource    = {dblp computer science bibliography, https://dblp.org}
}

@inproceedings{jelassi2024repeat,
  author       = {Samy Jelassi and
                  David Brandfonbrener and
                  Sham M. Kakade and
                  Eran Malach},
  title        = {Repeat After Me: Transformers are Better than State Space Models at
                  Copying},
  booktitle    = {Forty-first International Conference on Machine Learning, {ICML} 2024,
                  Vienna, Austria, July 21-27, 2024},
  publisher    = {OpenReview.net},
  year         = {2024},
  url          = {https://openreview.net/forum?id=duRRoGeoQT},
  bibsource    = {dblp computer science bibliography, https://dblp.org}
}

@article{trockman2024mimeticinitializationhelpsstate,
      title={Mimetic Initialization Helps State Space Models Learn to Recall}, 
      author={Asher Trockman and Hrayr Harutyunyan and J. Zico Kolter and Sanjiv Kumar and Srinadh Bhojanapalli},
      year={2024},
      journal={arXiv:2410.11135},
      archivePrefix={arXiv},
      primaryClass={cs.LG},
      url={https://arxiv.org/abs/2410.11135}, 
}

@inproceedings{okpekperevisiting,
  title={Revisiting Associative Recall in Modern Recurrent Models},
  author={Okpekpe, Destiny and Orvieto, Antonio},
  booktitle={First Workshop on Scalable Optimization for Efficient and Adaptive Foundation Models},
  year={2025},
  url={https://openreview.net/pdf?id=CcqAd5RPk5}
}

@inproceedings{poli2023hyena,
  author       = {Michael Poli and
                  Stefano Massaroli and
                  Eric Nguyen and
                  Daniel Y. Fu and
                  Tri Dao and
                  Stephen Baccus and
                  Yoshua Bengio and
                  Stefano Ermon and
                  Christopher R{\'{e}}},
  editor       = {Andreas Krause and
                  Emma Brunskill and
                  Kyunghyun Cho and
                  Barbara Engelhardt and
                  Sivan Sabato and
                  Jonathan Scarlett},
  title        = {Hyena Hierarchy: Towards Larger Convolutional Language Models},
  booktitle    = {International Conference on Machine Learning, {ICML} 2023, 23-29 July
                  2023, Honolulu, Hawaii, {USA}},
  series       = {Proceedings of Machine Learning Research},
  volume       = {202},
  pages        = {28043--28078},
  publisher    = {{PMLR}},
  year         = {2023},
  url          = {https://proceedings.mlr.press/v202/poli23a.html},
  bibsource    = {dblp computer science bibliography, https://dblp.org}
}

@inproceedings{fu2023hungry,
  author       = {Daniel Y. Fu and
                  Tri Dao and
                  Khaled Kamal Saab and
                  Armin W. Thomas and
                  Atri Rudra and
                  Christopher R{\'{e}}},
  title        = {Hungry Hungry Hippos: Towards Language Modeling with State Space Models},
  booktitle    = {The Eleventh International Conference on Learning Representations,
                  {ICLR} 2023, Kigali, Rwanda, May 1-5, 2023},
  publisher    = {OpenReview.net},
  year         = {2023},
  url          = {https://openreview.net/forum?id=COZDy0WYGg},
  bibsource    = {dblp computer science bibliography, https://dblp.org}
}

@article{liu2024longhornstatespacemodels,
      title={Longhorn: State Space Models are Amortized Online Learners}, 
      author={Bo Liu and Rui Wang and Lemeng Wu and Yihao Feng and Peter Stone and Qiang Liu},
      year={2024},
      journal={arXiv:2407.14207},
      archivePrefix={arXiv},
      primaryClass={cs.LG},
      url={https://arxiv.org/abs/2407.14207}, 
}

@inproceedings{lutati-etal-2023-focus,
    title = "Focus Your Attention (with Adaptive {IIR} Filters)",
    author = "Lutati, Shahar  and
      Zimerman, Itamar  and
      Wolf, Lior",
    editor = "Bouamor, Houda  and
      Pino, Juan  and
      Bali, Kalika",
    booktitle = "Proceedings of the 2023 Conference on Empirical Methods in Natural Language Processing",
    month = dec,
    year = "2023",
    address = "Singapore",
    publisher = "Association for Computational Linguistics",
    url = "https://aclanthology.org/2023.emnlp-main.772/",
    doi = "10.18653/v1/2023.emnlp-main.772",
    pages = "12538--12549",
}

@inproceedings{li2025cat,
  author       = {Mingchen Li and
                  Xuechen Zhang and
                  Yixiao Huang and
                  Samet Oymak},
  editor       = {Toby Walsh and
                  Julie Shah and
                  Zico Kolter},
  title        = {On the Power of Convolution-Augmented Transformer},
  booktitle    = {AAAI-25, Sponsored by the Association for the Advancement of Artificial
                  Intelligence, February 25 - March 4, 2025, Philadelphia, PA, {USA}},
  pages        = {18393--18402},
  publisher    = {{AAAI} Press},
  year         = {2025},
  url          = {https://doi.org/10.1609/aaai.v39i17.34024},
  doi          = {10.1609/AAAI.V39I17.34024},
  bibsource    = {dblp computer science bibliography, https://dblp.org}
}

@article{wang2025testtimeregressionunifyingframework,
      title={Test-time regression: a unifying framework for designing sequence models with associative memory}, 
      author={Ke Alexander Wang and Jiaxin Shi and Emily B. Fox},
      year={2025},
      journal={arXiv:2501.12352},
      archivePrefix={arXiv},
      primaryClass={cs.LG},
      url={https://arxiv.org/abs/2501.12352}, 
}

@article{elhage2021mathematical,
   title={A Mathematical Framework for Transformer Circuits},
   author={Elhage, Nelson and Nanda, Neel and Olsson, Catherine and Henighan, Tom and Joseph, Nicholas and Mann, Ben and Askell, Amanda and Bai, Yuntao and Chen, Anna and Conerly, Tom and DasSarma, Nova and Drain, Dawn and Ganguli, Deep and Hatfield-Dodds, Zac and Hernandez, Danny and Jones, Andy and Kernion, Jackson and Lovitt, Liane and Ndousse, Kamal and Amodei, Dario and Brown, Tom and Clark, Jack and Kaplan, Jared and McCandlish, Sam and Olah, Chris},
   year={2021},
   journal={Transformer Circuits Thread},
   url={https://transformer-circuits.pub/2021/framework/index.html}
}

@article{gu2024mambalineartimesequencemodeling,
      title={Mamba: Linear-Time Sequence Modeling with Selective State Spaces}, 
      author={Albert Gu and Tri Dao},
      year={2024},
      journal={arXiv:2312.00752},
      archivePrefix={arXiv},
      primaryClass={cs.LG},
      url={https://arxiv.org/abs/2312.00752}, 
}

@article{dao2024transformersssmsgeneralizedmodels,
      title={Transformers are {SSM}s: Generalized Models and Efficient Algorithms Through Structured State Space Duality}, 
      author={Tri Dao and Albert Gu},
      year={2024},
      journal={arXiv:2405.21060},
      archivePrefix={arXiv},
      primaryClass={cs.LG},
      url={https://arxiv.org/abs/2405.21060}, 
}

@inproceedings{feng2024bind,
  author       = {Jiahai Feng and
                  Jacob Steinhardt},
  title        = {How do Language Models Bind Entities in Context?},
  booktitle    = {The Twelfth International Conference on Learning Representations,
                  {ICLR} 2024, Vienna, Austria, May 7-11, 2024},
  publisher    = {OpenReview.net},
  year         = {2024},
  url          = {https://openreview.net/forum?id=zb3b6oKO77},
  bibsource    = {dblp computer science bibliography, https://dblp.org}
}

@inproceedings{kim-schuster-2023-entity,
    title = "Entity Tracking in Language Models",
    author = "Kim, Najoung  and
      Schuster, Sebastian",
    editor = "Rogers, Anna  and
      Boyd-Graber, Jordan  and
      Okazaki, Naoaki",
    booktitle = "Proceedings of the 61st Annual Meeting of the Association for Computational Linguistics (Volume 1: Long Papers)",
    month = jul,
    year = "2023",
    address = "Toronto, Canada",
    publisher = "Association for Computational Linguistics",
    url = "https://aclanthology.org/2023.acl-long.213/",
    doi = "10.18653/v1/2023.acl-long.213",
    pages = "3835--3855",
}

@article{li2025howlanguagemodelstrack,
      title={({H}ow) Do Language Models Track State?}, 
      author={Belinda Z. Li and Zifan Carl Guo and Jacob Andreas},
      year={2025},
      journal={arXiv:2503.02854},
      archivePrefix={arXiv},
      primaryClass={cs.CL},
      url={https://arxiv.org/abs/2503.02854}, 
}

@article{zeyuan,
    title={Physics of {L}anguage {M}odels: Part 4.1,
{A}rchitecture Design and the Magic of {C}anon Layers},
    author={Zeyuan Allen-Zhu and Alberto Alfarano},
    year={2025},
    journal={SSRN},
    url={https://papers.ssrn.com/sol3/papers.cfm?abstract_id=5240330},
}

@article{bick2025understandingskillgaprecurrent,
      title={Understanding the Skill Gap in Recurrent Language Models: The Role of the Gather-and-Aggregate Mechanism}, 
      author={Aviv Bick and Eric Xing and Albert Gu},
      year={2025},
      journal={arXiv:2504.18574},
      archivePrefix={arXiv},
      primaryClass={cs.LG},
      url={https://arxiv.org/abs/2504.18574}, 
}

@article{wen2024rnnstransformersyetkey,
      title={{RNN}s are not Transformers (Yet): The Key Bottleneck on In-context Retrieval}, 
      author={Kaiyue Wen and Xingyu Dang and Kaifeng Lyu},
      year={2024},
      journal={arXiv:2402.18510},
      archivePrefix={arXiv},
      primaryClass={cs.LG},
      url={https://arxiv.org/abs/2402.18510}, 
}

@article{waleffe2024empiricalstudymambabasedlanguage,
      title={An Empirical Study of {M}amba-based Language Models}, 
      author={Roger Waleffe and Wonmin Byeon and Duncan Riach and Brandon Norick and Vijay Korthikanti and Tri Dao and Albert Gu and Ali Hatamizadeh and Sudhakar Singh and Deepak Narayanan and Garvit Kulshreshtha and Vartika Singh and Jared Casper and Jan Kautz and Mohammad Shoeybi and Bryan Catanzaro},
      year={2024},
      journal={arXiv:2406.07887},
      archivePrefix={arXiv},
      primaryClass={cs.LG},
      url={https://arxiv.org/abs/2406.07887}, 
}

@inproceedings{causalabstraction,
 author = {Geiger, Atticus and Lu, Hanson and Icard, Thomas and Potts, Christopher},
 booktitle = {Advances in Neural Information Processing Systems},
 editor = {M. Ranzato and A. Beygelzimer and Y. Dauphin and P.S. Liang and J. Wortman Vaughan},
 pages = {9574--9586},
 publisher = {Curran Associates, Inc.},
 title = {Causal Abstractions of Neural Networks},
 url = {https://proceedings.neurips.cc/paper_files/paper/2021/file/4f5c422f4d49a5a807eda27434231040-Paper.pdf},
 volume = {34},
 year = {2021}
}

@article{geiger2024causalabstractiontheoreticalfoundation,
      title={Causal Abstraction: A Theoretical Foundation for Mechanistic Interpretability}, 
      author={Atticus Geiger and Duligur Ibeling and Amir Zur and Maheep Chaudhary and Sonakshi Chauhan and Jing Huang and Aryaman Arora and Zhengxuan Wu and Noah Goodman and Christopher Potts and Thomas Icard},
      year={2024},
      journal={arXiv:2301.04709},
      archivePrefix={arXiv},
      primaryClass={cs.AI},
      url={https://arxiv.org/abs/2301.04709}, 
}

@inproceedings{nanda2023progress,
  author       = {Neel Nanda and
                  Lawrence Chan and
                  Tom Lieberum and
                  Jess Smith and
                  Jacob Steinhardt},
  title        = {Progress measures for grokking via mechanistic interpretability},
  booktitle    = {The Eleventh International Conference on Learning Representations,
                  {ICLR} 2023},
  address = {Kigali, Rwanda},
  publisher    = {OpenReview.net},
  year         = {2023},
  url          = {https://openreview.net/forum?id=9XFSbDPmdW},
  bibsource    = {dblp computer science bibliography, https://dblp.org}
}

@inproceedings{singh2024what,
  author       = {Aaditya K. Singh and
                  Ted Moskovitz and
                  Felix Hill and
                  Stephanie C. Y. Chan and
                  Andrew M. Saxe},
  title        = {What needs to go right for an induction head? {A} mechanistic study
                  of in-context learning circuits and their formation},
  booktitle    = {Forty-first International Conference on Machine Learning, {ICML} 2024},
  address      = {Vienna, Austria},
  publisher    = {OpenReview.net},
  year         = {2024},
  url          = {https://openreview.net/forum?id=O8rrXl71D5},
  bibsource    = {dblp computer science bibliography, https://dblp.org}
}

@inproceedings{reddy2024mechanistic,
  author       = {Gautam Reddy},
  title        = {The mechanistic basis of data dependence and abrupt learning in an
                  in-context classification task},
  booktitle    = {The Twelfth International Conference on Learning Representations,
                  {ICLR} 2024},
  address      = {Vienna, Austria},
  publisher    = {OpenReview.net},
  year         = {2024},
  url          = {https://openreview.net/forum?id=aN4Jf6Cx69},
  bibsource    = {dblp computer science bibliography, https://dblp.org}
}

@inproceedings{tigges2024llm,
  author       = {Curt Tigges and
                  Michael Hanna and
                  Qinan Yu and
                  Stella Biderman},
  editor       = {Amir Globersons and
                  Lester Mackey and
                  Danielle Belgrave and
                  Angela Fan and
                  Ulrich Paquet and
                  Jakub M. Tomczak and
                  Cheng Zhang},
  title        = {{LLM} Circuit Analyses Are Consistent Across Training and Scale},
  booktitle    = {Advances in Neural Information Processing Systems 38: Annual Conference
                  on Neural Information Processing Systems 2024, NeurIPS 2024},
  address = {Vancouver, BC, Canada},
  year         = {2024},
  url          = {http://papers.nips.cc/paper\_files/paper/2024/hash/47c7edadfee365b394b2a3bd416048da-Abstract-Conference.html},
  bibsource    = {dblp computer science bibliography, https://dblp.org}
}

@article{yin2025attentionheadsmatterincontext,
      title={Which Attention Heads Matter for In-Context Learning?}, 
      author={Kayo Yin and Jacob Steinhardt},
      year={2025},
      journal={arXiv:2502.14010},
      archivePrefix={arXiv},
      primaryClass={cs.LG},
      url={https://arxiv.org/abs/2502.14010}, 
}

@inproceedings{edelman2024evolution,
  author       = {Ezra Edelman and
                  Nikolaos Tsilivis and
                  Benjamin L. Edelman and
                  Eran Malach and
                  Surbhi Goel},
  editor       = {Amir Globersons and
                  Lester Mackey and
                  Danielle Belgrave and
                  Angela Fan and
                  Ulrich Paquet and
                  Jakub M. Tomczak and
                  Cheng Zhang},
  title        = {The Evolution of Statistical Induction Heads: In-Context Learning
                  Markov Chains},
  booktitle    = {Advances in Neural Information Processing Systems 38: Annual Conference
                  on Neural Information Processing Systems 2024, NeurIPS 2024},
  address = {Vancouver, BC, Canada},
  year         = {2024},
  url          = {http://papers.nips.cc/paper\_files/paper/2024/hash/75b0edb869e2cd509d64d0e8ff446bc1-Abstract-Conference.html},
  bibsource    = {dblp computer science bibliography, https://dblp.org}
}

@inproceedings{nowak-cotterell-2023-fast,
    title = "A Fast Algorithm for Computing Prefix Probabilities",
    author = "Nowak, Franz  and
      Cotterell, Ryan",
    editor = "Rogers, Anna  and
      Boyd-Graber, Jordan  and
      Okazaki, Naoaki",
    booktitle = "Proceedings of the 61st Annual Meeting of the Association for Computational Linguistics (Volume 2: Short Papers)",
    month = jul,
    year = "2023",
    address = "Toronto, Canada",
    publisher = "Association for Computational Linguistics",
    url = "https://aclanthology.org/2023.acl-short.6/",
    doi = "10.18653/v1/2023.acl-short.6",
    pages = "57--69",
}

@inproceedings{wu-etal-2024-pyvene,
    title = "pyvene: A Library for Understanding and Improving {P}y{T}orch Models via Interventions",
    author = "Wu, Zhengxuan  and
      Geiger, Atticus  and
      Arora, Aryaman  and
      Huang, Jing  and
      Wang, Zheng  and
      Goodman, Noah  and
      Manning, Christopher  and
      Potts, Christopher",
    editor = "Chang, Kai-Wei  and
      Lee, Annie  and
      Rajani, Nazneen",
    booktitle = "Proceedings of the 2024 Conference of the North American Chapter of the Association for Computational Linguistics: Human Language Technologies (Volume 3: System Demonstrations)",
    month = jun,
    year = "2024",
    address = "Mexico City, Mexico",
    publisher = "Association for Computational Linguistics",
    url = "https://aclanthology.org/2024.naacl-demo.16/",
    doi = "10.18653/v1/2024.naacl-demo.16",
    pages = "158--165",
}

@inproceedings{attention,
  author       = {Ashish Vaswani and
                  Noam Shazeer and
                  Niki Parmar and
                  Jakob Uszkoreit and
                  Llion Jones and
                  Aidan N. Gomez and
                  Lukasz Kaiser and
                  Illia Polosukhin},
  editor       = {Isabelle Guyon and
                  Ulrike von Luxburg and
                  Samy Bengio and
                  Hanna M. Wallach and
                  Rob Fergus and
                  S. V. N. Vishwanathan and
                  Roman Garnett},
  title        = {Attention is All you Need},
  booktitle    = {Advances in Neural Information Processing Systems 30: Annual Conference
                  on Neural Information Processing Systems 2017},
  address = {Long Beach, CA, {USA}},
  pages        = {5998--6008},
  year         = {2017},
  url          = {https://proceedings.neurips.cc/paper/2017/hash/3f5ee243547dee91fbd053c1c4a845aa-Abstract.html},
  bibsource    = {dblp computer science bibliography, https://dblp.org}
}

@inproceedings{prakash2024binding,
  author       = {Nikhil Prakash and
                  Tamar Rott Shaham and
                  Tal Haklay and
                  Yonatan Belinkov and
                  David Bau},
  title        = {Fine-Tuning Enhances Existing Mechanisms: {A} Case Study on Entity
                  Tracking},
  booktitle    = {The Twelfth International Conference on Learning Representations,
                  {ICLR} 2024},
  address = {Vienna, Austria},
  publisher    = {OpenReview.net},
  year         = {2024},
  url          = {https://openreview.net/forum?id=8sKcAWOf2D},
  bibsource    = {dblp computer science bibliography, https://dblp.org}
}

@inproceedings{wang2024grok,
  author       = {Boshi Wang and
                  Xiang Yue and
                  Yu Su and
                  Huan Sun},
  editor       = {Amir Globersons and
                  Lester Mackey and
                  Danielle Belgrave and
                  Angela Fan and
                  Ulrich Paquet and
                  Jakub M. Tomczak and
                  Cheng Zhang},
  title        = {Grokking of Implicit Reasoning in Transformers: {A} Mechanistic Journey
                  to the Edge of Generalization},
  booktitle    = {Advances in Neural Information Processing Systems 38: Annual Conference
                  on Neural Information Processing Systems 2024, NeurIPS 2024, Vancouver,
                  BC, Canada, December 10 - 15, 2024},
  year         = {2024},
  url          = {http://papers.nips.cc/paper\_files/paper/2024/hash/ad217e0c7fecc71bdf48660ad6714b07-Abstract-Conference.html},
  bibsource    = {dblp computer science bibliography, https://dblp.org}
}

@inproceedings{brinkmann-etal-2024-mechanistic,
    title = "A Mechanistic Analysis of a Transformer Trained on a Symbolic Multi-Step Reasoning Task",
    author = "Brinkmann, Jannik  and
      Sheshadri, Abhay  and
      Levoso, Victor  and
      Swoboda, Paul  and
      Bartelt, Christian",
    editor = "Ku, Lun-Wei  and
      Martins, Andre  and
      Srikumar, Vivek",
    booktitle = "Findings of the Association for Computational Linguistics: ACL 2024",
    month = aug,
    year = "2024",
    address = "Bangkok, Thailand",
    publisher = "Association for Computational Linguistics",
    url = "https://aclanthology.org/2024.findings-acl.242/",
    doi = "10.18653/v1/2024.findings-acl.242",
    pages = "4082--4102"
}

@inproceedings{wiegreffe2025answer,
title={Answer, Assemble, Ace: Understanding How {LM}s Answer Multiple Choice Questions},
author={Sarah Wiegreffe and Oyvind Tafjord and Yonatan Belinkov and Hannaneh Hajishirzi and Ashish Sabharwal},
booktitle={The Thirteenth International Conference on Learning Representations},
year={2025},
url={https://openreview.net/forum?id=6NNA0MxhCH}
}

@article{lieberum2023does,
  title={Does circuit analysis interpretability scale? evidence from multiple choice capabilities in chinchilla},
  author={Lieberum, Tom and Rahtz, Matthew and Kram{\'a}r, J{\'a}nos and Nanda, Neel and Irving, Geoffrey and Shah, Rohin and Mikulik, Vladimir},
  journal={arXiv preprint arXiv:2307.09458},
  year={2023}
}

@inproceedings{liu2024short,
  author       = {Zicheng Liu and
                  Siyuan Li and
                  Li Wang and
                  Zedong Wang and
                  Yunfan Liu and
                  Stan Z. Li},
  title        = {Short-Long Convolutions Help Hardware-Efficient Linear Attention to
                  Focus on Long Sequences},
  booktitle    = {Forty-first International Conference on Machine Learning, {ICML} 2024},
  address = {Vienna, Austria},
  publisher    = {OpenReview.net},
  year         = {2024},
  url          = {https://openreview.net/forum?id=TRrXkVdhwi},
  bibsource    = {dblp computer science bibliography, https://dblp.org}
}

@article{dolga2024lattelatentattentionlinear,
      title={Latte: Latent Attention for Linear Time Transformers}, 
      author={Rares Dolga and Lucas Maystre and Marius Cobzarenco and David Barber},
      year={2024},
      journal={arXiv:2402.17512},
      archivePrefix={arXiv},
      primaryClass={cs.CL},
      url={https://arxiv.org/abs/2402.17512}, 
}

@article{greff2020bindingproblemartificialneural,
      title={On the Binding Problem in Artificial Neural Networks}, 
      author={Klaus Greff and Sjoerd van Steenkiste and Jürgen Schmidhuber},
      year={2020},
      journal={arXiv:2012.05208},
      archivePrefix={arXiv},
      primaryClass={cs.NE},
      url={https://arxiv.org/abs/2012.05208}, 
}

@article{prakash2025languagemodelsuselookbacks,
      title={Language Models use Lookbacks to Track Beliefs}, 
      author={Nikhil Prakash and Natalie Shapira and Arnab Sen Sharma and Christoph Riedl and Yonatan Belinkov and Tamar Rott Shaham and David Bau and Atticus Geiger},
      year={2025},
      journal={arXiv:2505.14685},
      archivePrefix={arXiv},
      primaryClass={cs.CL},
      url={https://arxiv.org/abs/2505.14685}, 
}

@software{yang2024fla,
  title  = {FLA: A Triton-Based Library for Hardware-Efficient Implementations of Linear Attention Mechanism},
  author = {Yang, Songlin and Zhang, Yu},
  url    = {https://github.com/fla-org/flash-linear-attention},
  month  = jan,
  year   = {2024}
}

@inproceedings{deltanet,
  author       = {Songlin Yang and
                  Bailin Wang and
                  Yu Zhang and
                  Yikang Shen and
                  Yoon Kim},
  editor       = {Amir Globersons and
                  Lester Mackey and
                  Danielle Belgrave and
                  Angela Fan and
                  Ulrich Paquet and
                  Jakub M. Tomczak and
                  Cheng Zhang},
  title        = {Parallelizing Linear Transformers with the Delta Rule over Sequence
                  Length},
  booktitle    = {Advances in Neural Information Processing Systems 38: Annual Conference
                  on Neural Information Processing Systems 2024, NeurIPS 2024},
  location = {Vancouver,
                  BC, Canada},
  year         = {2024},
  url          = {http://papers.nips.cc/paper\_files/paper/2024/hash/d13a3eae72366e61dfdc7eea82eeb685-Abstract-Conference.html},
  bibsource    = {dblp computer science bibliography, https://dblp.org}
}

@article{shazeer2020gluvariantsimprovetransformer,
      title={GLU Variants Improve Transformer}, 
      author={Noam Shazeer},
      year={2020},
      journal={arXiv:2002.05202},
      archivePrefix={arXiv},
      primaryClass={cs.LG},
      url={https://arxiv.org/abs/2002.05202}, 
}
\bibliographystyle{iclr2026/iclr2026_conference}

%%%%%%%%%%%%%%%%%%%%%%%%%%%%%%%%%%%%%%%%%%%%%%%%%%%%%%%%%%%%

\newpage
\appendix
\addcontentsline{toc}{section}{Appendix} % Add the appendix text to the document TOC
\noptcrule
\part{Appendix} % Start the appendix part
\parttoc % Insert the appendix TOC

\newpage
\section{Model configurations}
\label{sec:config}

\begin{table}[!h]
\centering
\caption{Default model configurations across all architectures. In experiments, we sweep learning rate and embedding dimension, reporting results from the instance with highest accuracy.}
\small
\begin{subtable}[t]{0.3\textwidth}
    \centering
    \caption{Attention}
    \begin{tabular}{lc}
    \toprule
    \textbf{Parameter} & \textbf{Values} \\
    \midrule
    \verb|dropout| & $0.0$ \\
    \verb|num_heads| & $1$ \\
    \bottomrule
    \end{tabular}
\end{subtable}
\hfill
\begin{subtable}[t]{0.3\textwidth}
    \centering
    \caption{Hyena}
    \begin{tabular}{lc}
    \toprule
    \textbf{Parameter} & \textbf{Values} \\
    \midrule
    \verb|l_max| & $1024$ \\
    \verb|filter_order| & $64$ \\
    \verb|num_heads| & $1$ \\
    \verb|num_blocks| & $1$ \\
    \verb|outer_mixing| & False \\
    \verb|dropout| & $0.0$ \\
    \verb|filter_dropout| & $0.0$ \\
    \verb|short_filter_order| & $3$ \\
    \verb|bidirectional| & False \\
    \bottomrule
    \end{tabular}
\end{subtable}
\hfill
\begin{subtable}[t]{0.3\textwidth}
    \centering
    \caption{BaseConv}
    \begin{tabular}{lc}
    \toprule
    \textbf{Parameter} & \textbf{Values} \\
    \midrule
    \verb|l_max| & $1024$ \\
    \verb|kernel_size| & $[3, -1]$ \\
    \verb|implicit_long_conv| & True \\
    \verb|use_act| & False \\
    \bottomrule
    \end{tabular}
\end{subtable}\\
\begin{subtable}[t]{0.4\textwidth}
    \centering
    \caption{Based}
    \begin{tabular}{lc}
    \toprule
    \textbf{Parameter} & \textbf{Values} \\
    \midrule
    \textit{BaseConv Layer} \\
    \verb|l_max| & $1024$ \\
    \verb|kernel_size| & $3$ \\
    \verb|implicit_long_conv| & True \\
    \verb|use_act| & False \\
    \midrule
    \textit{Based Layer} \\
    \verb|l_max| & $1024$ \\
    \verb|feature_dim| & 8 \\
    \verb|num_key_value_heads| & 1 \\
    \verb|num_heads| & 1 \\
    \verb|feature_name| & \verb|taylor_exp| \\
    \verb|train_view| & \verb|quadratic| \\
    \bottomrule
    \end{tabular}
\end{subtable}
\hfill
\begin{subtable}[t]{0.28\textwidth}
    \centering
    \caption{H3}
    \begin{tabular}{lc}
    \toprule
    \textbf{Parameter} & \textbf{Values} \\
    \midrule
    \verb|l_max| & $1024$ \\
    \verb|d_state| & $1024$ \\
    \verb|head_dim| & $1024$ \\
    \bottomrule
    \end{tabular}
\end{subtable}
\hfill
\begin{subtable}[t]{0.28\textwidth}
    \centering
    \caption{Mamba}
    \begin{tabular}{lc}
    \toprule
    \textbf{Parameter} & \textbf{Values} \\
    \midrule
    \verb|d_conv| & $4$ \\
    % \verb|d_state| & $1024$ \\
    % \verb|head_dim| & $1024$ \\
    \bottomrule
    \end{tabular}
\end{subtable}
\end{table}

\newpage
% \section{ATR: a new language-like synthetic recall task}

% To deepen our findings, we introduce \textbf{\ourname{}} (\ourshort{}), a novel synthetic retrieval task more similar to real-world natural language retrieval than AR. \ourshort{} uses a probabilistic context-free grammar (PCFG) to generate hierarchical data, on which we ask AR-like queries. Since keys and values need not be adjacent to each other, \ourshort{} requires a true non-positional retrieval mechanism, which may challenge architectures that are designed for AR. Interestingly, we observe the same mechanisms are implicated across architectures on ATR as on AR, indicating that association mechanisms are not task-dependent.

\section{Formalisation of ATR}
\label{sec:formal}

\begin{table}[t!]
    \small
    \centering
    \begin{tabular}{cl}
        \toprule
        \textbf{Param.} & \textbf{Description} \\
        \midrule
        $H$ & Is the head terminal at the left or the right of each production? \\
        $d_{\textrm{max}}$ & Maximum depth permitted for the PCFG to generate.
        \\
        $L_\textrm{max}$ & Maximum number of symbols of the right-hand side of a production rule. \\
        $R_\textrm{max}$ & Maximum number of production rules for each nonterminal. \\
        $\lvert \mathcal{N} \rvert$ & Number of nonterminal symbols in the PCFG vocabulary. \\
        $\lvert \Sigma \rvert$ & Number of terminal symbols in the PCFG vocabulary. \\
        $r_{\Sigma}$ & Relative weightage on choosing a terminal when sampling production rules. \\
        \bottomrule
    \end{tabular}
    \caption{Parameters used for constructing a PCFG. We define PCFGs in Greibach Normal Form (GNF); see Appendix~\ref{sec:formal} for more details.}
    \label{tab:pcfg-params}
\end{table}

For reference, we provide formal definitions for PCFGs and the normal form we use in \ourshort{}.\footnote{We use similar formalisations of PCFGs as previous work in NLP, e.g.~\citet{nowak-cotterell-2023-fast}.}

% \aar{is any of this formalisation necessary? we def need a figure explaining what a PCFG is regardless}

\begin{definition}
A \textbf{probabilistic context-free grammar} is a tuple $\mathcal{G} = \langle \mathcal{N}, \Sigma, \mathrm{S}, \mathcal{R}, p\rangle$ where:
\begin{itemize}[noitemsep]
    \item $\mathcal{N}$ is a finite set of non-terminal symbols;
    \item $\Sigma$ is an alphabet of terminal symbols;
    \item $\mathrm{S} \in \mathcal{N}$ is a start symbol;
    % \item $h : \mathcal{N} \times \Sigma \to [0,1]$ is a weighting function which assigns a probability indicating how likely a terminal $a \in \Sigma$ is to be the head of a nonterminal $\mathrm{X} \in \mathcal{N}$, such that $ \sum_{a \in \Sigma} h(\mathrm{X}, a) = 1$ for all $x \in \mathcal{N}$;
    \item $\mathcal{R} \subset \mathcal{N} \times (\mathcal{N} \cup \Sigma)^*$ is a finite set of production rules, mapping a left-hand side symbol $\mathrm{N} \in \mathcal{N}$ to a string of symbols that may be either terminals or nonterminals; each such rule is written as $\mathrm{X} \to \bm{\alpha}$;
    \item $p : \mathcal{R} \to [0, 1]$ is a weighting function which assigns a probability to each production rule for a nonterminal; this function is locally normalised, meaning $\{ \sum_{\mathrm{X} \to \bm{\alpha}}{p(\mathrm{X} \to \bm{\alpha})} = 1 \mid \mathrm{X} \in \mathcal{N}\}$.
\end{itemize}
\end{definition}

\begin{definition} A PCFG $\mathcal{G} = \langle \mathcal{N}, \Sigma, \mathrm{S}, \mathcal{R}, p\rangle$ is in \textbf{Greibach normal form (GNF)} if each production rule in $\mathcal{R}$ is of the form $\mathrm{X} \to a~\mathrm{X_1}~\ldots~\mathrm{X_n}$,
where $\mathrm{X}_1, \ldots, \mathrm{X}_n \in \mathcal{N}$ and $n$ may be $0$. Similarly, a PCFG is in \textbf{right-Greibach normal form} if each rule is of the form $\mathrm{X} \to \mathrm{X_1}~\ldots~\mathrm{X_n}~a$.
\end{definition}

For \ourshort{}, the PCFG is in Greibach normal form if the head is the leftmost symbol of the production rule's righthand side; similarly, if the PCFG is right-headed, it is in right-Greibach normal form.

\begin{definition}
A \textbf{derivation step} $\bm{\alpha} \Rightarrow \bm{\beta}$ is an operation where, given strings of symbols $\bm{\alpha}, \bm{\beta} \in (\mathcal{N} \cup \Sigma)^*$, the leftmost nonterminal $\mathrm{X} \in \mathcal{N}$ in $\bm{\alpha}$ is rewritten using the right-hand side of a production rule $\mathrm{X} \to \ldots \in \mathcal{R}$ to obtain $\bm{\beta}$.

% Repeatedly applying the operation $\Rightarrow$ given the start symbol and sampling production rules from $\mathcal{R}$ per weights $p$ results in a sequence of terminals, which is a \textbf{sample} from the PCFG $\mathcal{G}$.
\end{definition}

\begin{definition}
A \textbf{derivation} under the PCFG $\mathcal{G}$ is a sequence of strings $[\bm{\alpha}_0, \ldots, \bm{\alpha}_m]$ where $\bm{\alpha}_0 \in \mathcal{N}$ and each step $\bm{\alpha}_{i + 1}$ is formed by a derivation step on $\bm{\alpha}_i$. The final string $\bm{\alpha}_m \in \Sigma^*$ is the \textbf{yield} of the derivation.
\end{definition}

Each ATR document is the yield of a derivation sampled under the GNF PCFG $\mathcal{G}$.

\subsection{Additional details on ATR}

\paragraph{Parent terminals in GNF.} We set the left/right-most terminal in each production rule (which leads to the GNF property) the parent of all other generated terminals. This terminal is sampled specially:~for each nonterminal, we independently sample a distribution over terminals from a uniform Dirichlet, and for all production rules with that nonterminal on the lefthand side we use that distribution to sample the parent terminal. This simulates how heads of phrases in natural language (analogous to our parent terminals) decide the type of the phrase they head (analogous to our nonterminals).

\paragraph{Maximum depth.} To enforce maximum depth, we first assign a uniformly random depth score $d : \mathcal{N} \to \mathbb{N} \in \{1, \ldots,  \texttt{max\_depth}\}$ to each nonterminal in the vocabulary. Then, for each production rule for each nonterminal $\mathrm{X}$, we only allow nonterminals $\mathrm{Y}$ with $d(\mathrm{Y}) > d(\mathrm{X})$ on the right-hand side. Note that this means no recursion is possible.

\section{Task hyperparameters}\label{sec:task-params}

\begin{table}[!h]
    \centering
    \caption{Task hyperparameters used for constructing key--value sets for AR and PCFGs for ATR. For a description of each parameter, see Table~\ref{tab:pcfg-params}.}
\begin{subtable}[t]{0.47\textwidth}
    \centering
    \caption{Parameters used for constructing AR documents.}
    \begin{tabular}{lc}
        \toprule
        \textbf{Parameter} & \textbf{Values} \\
        \midrule
        $L_\textrm{max}$ & $32$ \\
        $L_\textrm{min}$ & $32$ \\
        $\lvert \Sigma \rvert$ & $\{8192\}$ \\
        \bottomrule
    \end{tabular}
    \label{tab:ar-hparams}
\end{subtable}
\hfill
\begin{subtable}[t]{0.47\textwidth}
    \centering
    \caption{Parameters used for constructing ATR documents.}
    \begin{tabular}{lc}
        \toprule
        \textbf{Parameter} & \textbf{Values} \\
        \midrule
        $H$ & Right \\
        $d_{\textrm{max}}$ & $10$
        %\footnote{To enforce maximum depth, we first assign a uniformly random depth score $d : \mathcal{N} \to \mathbb{N} \in \{1, \ldots,  \texttt{max\_depth}\}$ to each nonterminal in the vocabulary. Then, for each production rule for each nonterminal $\mathrm{X}$, we only allow nonterminals $\mathrm{Y}$ with $d(\mathrm{Y}) > d(\mathrm{X})$ on the right-hand side. Note that this means no recursion is possible.} 
        \\
        $L_\textrm{max}$ & $5$ \\
        $R_\textrm{max}$ & $5$ \\
        $\lvert \mathcal{N} \rvert$ & $40$ \\
        $\lvert \Sigma \rvert$ & $20$ \\
        $r_{\Sigma}$ & $20$ \\
        \bottomrule
    \end{tabular}
    \label{tab:pcfg-hparams}
\end{subtable}
\end{table}

\newpage
\section{Additional experiments on AR and ATR}\label{sec:additional-exps}

Many parameters of synthetic tasks like AR and ATR and the model architectures we tested have interesting effects on behavioural and mechanistic metrics, but not all experiments could fit in our main text. Therefore, we include additional interesting observations in this appendix.

\subsection{Attention needs position embeddings}
\begin{figure}[!h]
    \centering
    \includegraphics[width=\linewidth]{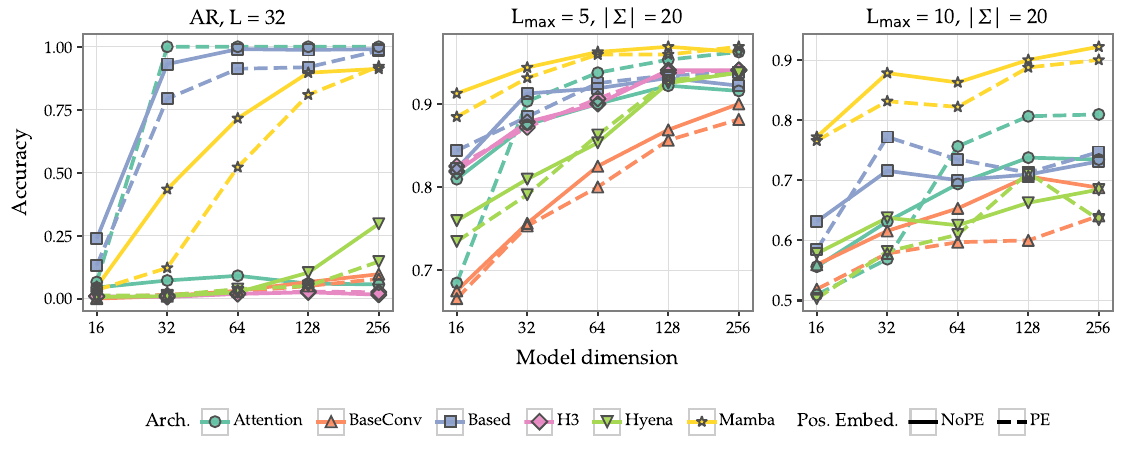}
    \caption{\textbf{Position embedding}: Model accuracy on AR and two ATR settings with and without absolute position embeddings.}
    \label{fig:pos}
\end{figure}

Due to an initial configuration mistake, we accidentally trained all architectures with absolute position embeddings; in the \texttt{zoology} codebase \citep{arora2024zoology}, only Attention is meant to be trained in this way. Fortuitously, this resulted in an interesting ablation: do SSMs, which are usually trained without it, also benefit from position embeddings?

\paragraph{Behavioural results.} Our results in \cref{fig:pos} resoundingly show no: SSMs generally perform worse with position embeddings (PE). Attention is highly dependent on PE; performance on AR drops from $100.00\%$ to $5.62\%$ at $d=256$ with NoPE. Attention lacks recurrence, unlike SSMs, so this is not surprising. However, on ATR, at smaller dimensionalities NoPE actually outperforms PE Attention. Further ablations ought to consider alternative PE methods such as RoPE and Alibi.

\subsection{Mamba's solution to ATR does generalise}
\label{sec:generalise}

We reuse the settings from our ATR experiment ($L = 5, \lvert \Sigma \rvert = 20$) and construct a new dataset with a train--test split on query--answer pairs. Specifically, $80\%$ of possible unique query--answer pairs are provided in the training set, while $20\%$ are only in the test set and thus never trained on. We seek to assess whether models learn a general mechanism for parent--child relations in ATR or if the impressive results of Mamba (as well as Attention and Based) are merely the result of better memorisation of the PCFG parameters.
This setup is akin to \citet{wang2024grok}'s technique of train--test split on multi-hop queries; we provide supervision on individual query and answer types, but not on some compositions of them.

\begin{figure}
    \centering
    \begin{subfigure}[t]{0.31\linewidth}
        \includegraphics[width=\linewidth]{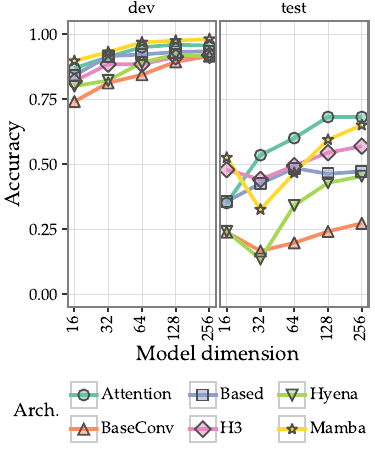}
        \caption{Model dimensionality vs.~accuracy on checkpoints with highest dev accuracy.}
        \label{fig:atr-gen-behaviour}
    \end{subfigure}
    \hfill
    \begin{subfigure}[t]{0.31\linewidth}
        \includegraphics[width=\linewidth]{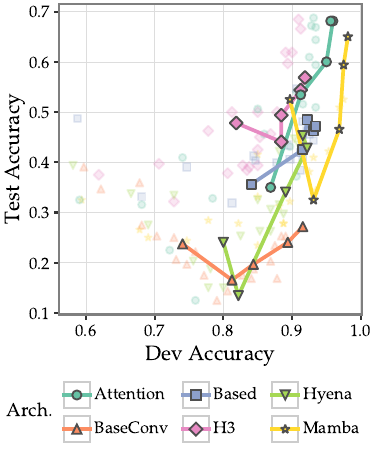}
        \caption{Dev vs.~test accuracy, with highest dev accuracy checkpoints at each dim.~highlighted.}
        \label{fig:atr-gen-comp}
    \end{subfigure}
    \hfill
    \begin{subfigure}[t]{0.31\linewidth}
        \includegraphics[width=\linewidth]{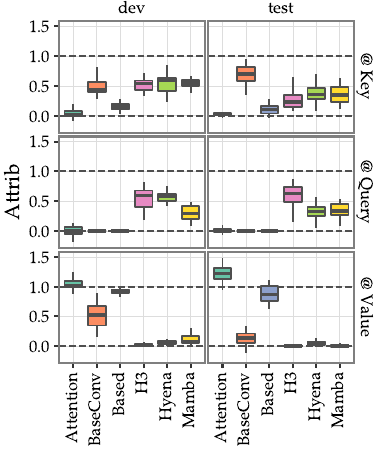}
        \caption{Attribution scores for all checkpoints (except outliers), compared between dev and test sets.}
        \label{fig:atr-gen-interp}
    \end{subfigure}
    \caption{\textbf{Generalisation on Associative Treecall}: Accuracy and interchange intervention results on ATR with train--test split. Scores are reported on dev (with in-distribution query--answer pairs from training) and test (OOD). We highlight the checkpoint with the best dev score in each setting.}
    \label{fig:atr-gen}
\end{figure}

\paragraph{Behavioural results.} We select the checkpoint with the highest dev accuracy for each architectural and dimensionality setting, after sweeping LR. We plot the dev and test accuracies of each of these checkpoints in \cref{fig:atr-gen-behaviour}; all models have much lower test accuracy (e.g.~Attention with $d=256$ has $95.62\%$ dev and $68.12\%$ test accuracy). Attention achieves the greatest dev accuracies on $d \geq 32$. Mamba's relative ranking is lower than on the in-distribution setting in \cref{sec:atr-exp}, but it still achieves the overall second-highest dev accuracy ($65.00\%$ at $d=128$). Surprisingly, H3 generalises well despite its poor dev accuracy, beating Mamba on test accuracy in $3$ out of $5$ settings.

We compare dev and test accuracies across all LRs in \cref{fig:atr-gen-comp}. We find that while Mamba does have unusually high dev accuracy given a selected test accuracy (indicating greater memorisation than models with other architectures), its dev accuracy is still generally higher than non-Attention architectures. Interestingly, H3 has nearly Attention-level generalisation while BaseConv exhibits vanishingly little generalisation. Overall, behavioural metrics show that Mamba does nontrivially generalise on ATR, albeit not as well as Attention.

\paragraph{Mechanistic analysis.}  We report a summary of attribution scores at different tokens (\textit{key}, \textit{query}, \textit{value}), comparing on dev and test sets across all checkpoints in \cref{fig:atr-gen-interp}. We find largely consistent mechanisms underlying behaviour on both dev and test, and these match attribution scores on ATR without train--test split. The only exception is that BasedConv does induction on the dev set but not nearly as much on the test set; its induction mechanism is more brittle than Attention and Based.

Overall, the induction mechanism is not more general than the direct retrieval mechanism; both Attention and Mamba show greater generalisation than other architectures despite their entirely different solutions, and our mechanistic evaluations confirm that this solution is consistent across in-distribution and out-of-distribution queries.

\subsection{1-layer SSMs learn direct retrieval on AR and ATR}
\begin{figure}[!h]
    \centering
    \begin{subfigure}[t]{0.48\linewidth}
        \includegraphics[width=\linewidth]{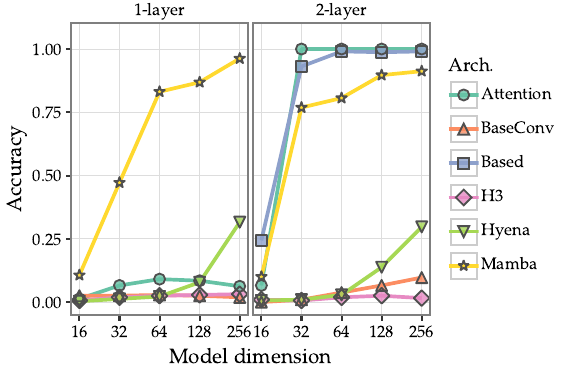}
        \caption{Accuracy of 1-layer vs.~2-layer models on AR, 32 key--value pairs. 1-layer induction models fail.}
        \label{fig:ar-layers}
    \end{subfigure}
    \hfill
    \begin{subfigure}[t]{0.48\linewidth}
        \includegraphics[width=\linewidth]{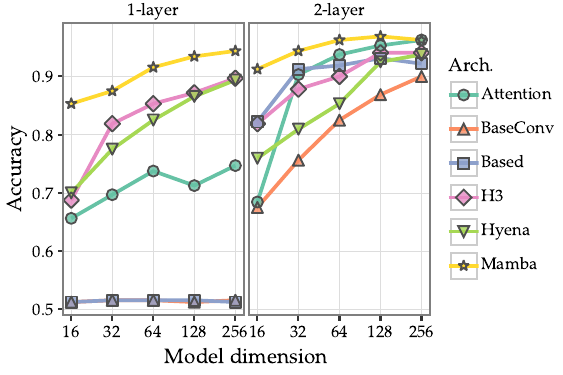}
        \caption{Accuracy of 1-layer vs.~2-layer models on ATR ($L=5, \lvert \Sigma \rvert = 20$), with Based and BaseConv failing.}
        \label{fig:atr-layers}
    \end{subfigure}
    \caption{\textbf{1-layer models on AR and ATR}: Architectures that learn induction in the 2-layer setting fail to perform non-trivially with 1 layer. Mamba is highly performant with 1 layer on both tasks.}
    \label{fig:enter-label}
\end{figure}

Throughout our experiments on AR and ATR, we have claimed that SSMs (except for Based and possibly BaseConv) learn a direct retrieval mechanism which does not require an intermediate step like attention, i.e.~only a single SSM layer is needed to learn AR and ATR. To verify this, we repeat AR and $L=5, \lvert \Sigma \rvert = 20$ ATR experiments (without train--test split) with 1-layer models.

\paragraph{Behavioural results.} We find comparable performance for direct retrieval models between 1-layer and 2-layer settings on AR (\cref{fig:ar-layers}). In fact, at $d=256$, 1-layer Mamba ($96.25\%$) outperforms 2-layer Mamba ($91.25\%$), as does Hyena ($31.56\%$ vs.~$29.69\%$). 1-layer Based and BaseConv are architecturally identical, so we only report one; that architecture and Attention, both relying on induction in the 2-layer case, fail to learn AR with one layer. On ATR (\cref{fig:atr-layers}), we see a more noticeable difference with layer count on all architectures, but again Attention, Based, and BaseConv become the worst architectures with one layer (e.g.~$96.25\% \to 74.69\%$ for Attention at $d=256$).

\subsection{SSMs prefer layer 0 to perform AR}

\begin{figure}[!h]
    \centering
    \begin{subfigure}[b]{0.48\linewidth}
        \includegraphics[width=\linewidth]{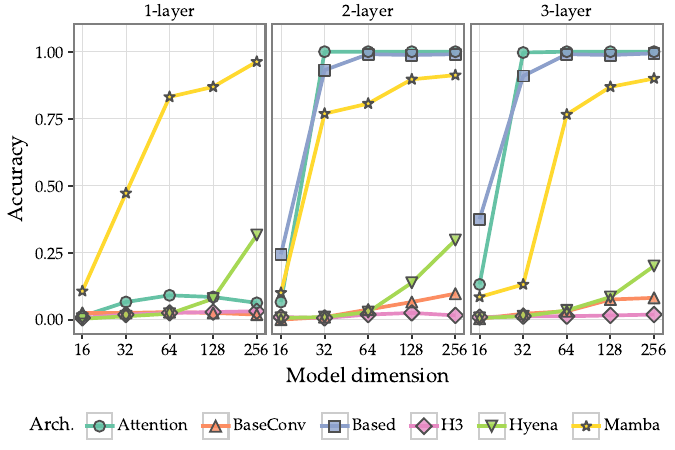}
        \caption{Accuracy with 1--3 layers on AR.}
        \label{fig:ar-three-acc}
    \end{subfigure}
    \hfill
    \begin{subfigure}[b]{0.48\linewidth}
        \includegraphics[width=\linewidth]{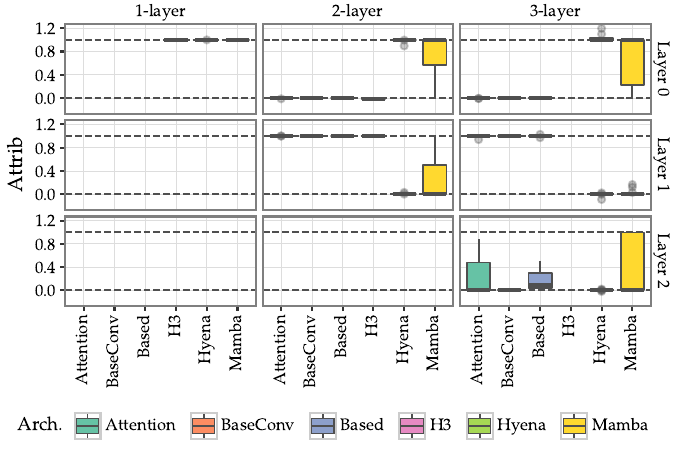}
        \caption{Attribution scores with 1--3 layers on AR.}
        \label{fig:ar-three-attrib}
    \end{subfigure}
    \caption{\textbf{Varying layer count on AR}: Behavioural and mechanistic evaluations for models with 1--3 layers on AR.}
    \label{fig:ar-three}
\end{figure}

Since we have confirmed that the direct retrieval mechanism in SSMs requires only a single layer, we are curious which layer this mechanism forms in if more than two layers are present. We train models with up to three layers on AR and report results.

\paragraph{Behavioural results.} 3-layer models perform about the same on AR as 2-layer models across architectures (\cref{fig:ar-three-acc}), except for a large drop in performance for Mamba when $d=32$; this may just be an optimisation failure.

\paragraph{Mechanistic analysis.} For our mechanistic metric, instead of intervening on each block, we intervene at the sequence mixer's output to the \textit{query} token in each layer; this tells us if that layer is directly responsible for writing the answer to the output position. We apply the same filter as in \cref{sec:ar-exp}, with a threshold of $0.01$. \Cref{fig:ar-three-attrib} shows that among performant models, Hyena and Mamba prefer layer 0 for performing AR no matter the layer count; however, some Mamba checkpoints learn the mechanism in the final layer as well (but never layer 1 in a 3-layer model). Attention, Based, and BaseConv prefer layer 1, which is expected since this is the second step of the induction mechanism. However, some checkpoints of Attention and Based also have non-zero attribution score at layer 2 in the 3-layer setting.

\subsection{Rightmost sibling queries are trivial for all architectures}

\begin{figure}[!h]
    \centering
    \includegraphics[width=0.8\linewidth]{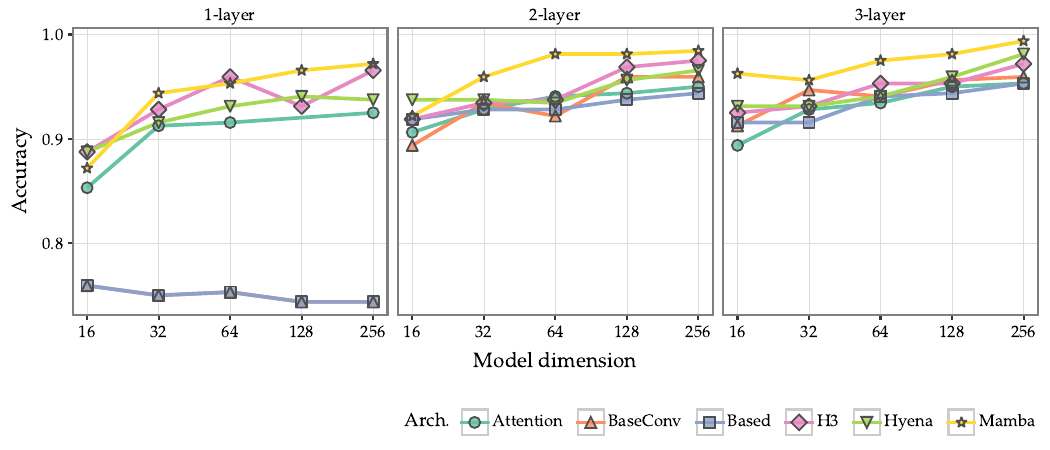}
    \caption{\textbf{Sibling queries}: Accuracy across models with 1--3 layers on ATR ($L_{\textrm{max}}=5, \lvert \Sigma \rvert = 20$)}.
    \label{fig:multihop}
\end{figure}

Since ATR has hierarchical structure, we attempted an initial experiment with multihop queries; specifically, we present queries where the answer is that terminal's rightmost sibling terminal. Models are only trained on this type of query, not standard parent queries as reported in the main text. We train with the same settings in \cref{sec:methodology}.

\paragraph{Behavioural results.} In \cref{fig:multihop} we show that all models (except Based and BasedConv with $1$ layer, where they only have local convolutions) achieve greater than $80\%$ accuracy at the task at all dimensionalities. We see slight improvement from 1-layer to 2-layer models but at this point performance is saturated and 3-layer does not help. Clearly, this task is extremely simple for all models, even more so than parent queries, and thus does not provide useful signal for comparing architectures.

\paragraph{Why are sibling queries easy?} Parent nodes are guaranteed to be special terminals in our GNF which are sampled from a nonterminal-dependent distribution (see \cref{sec:formal}). However, siblings have a large chance of being fixed terminals specified by the production rule. Additionally, the rightmost sibling of a particular terminal may be itself, if it is the rightmost terminal of its production rule. We speculate that these factors combined make sibling queries easier than parent queries, and thus not a suitable testbed for multihop reasoning.

\paragraph{Future work.} The appropriate analogue to study multihop \textit{reasoning} in ATR is grandparent relations (or higher up ancestors in the tree), since the grandparent is always a special head terminal (like the parent) and is always to the right of the parent and thus different from the query terminal. We leave further experiments on this to future work.

%%%%%%%%%%%%%%%%%%%%%%%%%%%%%%%%%%%%%%%%%%%%%%%%%%%%%%%%%%%%

\newpage

\end{document}